\documentclass{article} 
\usepackage[final]{colm2026_conference}

\usepackage{multicol}
\usepackage{microtype}
\usepackage{hyperref}
\usepackage{url}
\usepackage{booktabs}
\usepackage{multirow}
\usepackage{makecell}
\usepackage{graphicx}
\usepackage{amsmath,amssymb}
\usepackage{enumitem}
\usepackage{tikz}
\usepackage{listings}
\usepackage{xcolor}
\usepackage[table]{xcolor}
\usepackage{lineno}
\usepackage{subcaption} 
\usepackage{placeins}

\usetikzlibrary{arrows.meta,positioning,calc}

\definecolor{darkblue}{rgb}{0, 0, 0.5}
\hypersetup{
  colorlinks=true,
  citecolor=darkblue,
  linkcolor=darkblue,
  urlcolor=darkblue
}

\lstdefinelanguage{JSON}{
  basicstyle=\ttfamily\small,
  breaklines=true,
  literate=
   *{0}{{{\color{blue}0}}}{1}
    {1}{{{\color{blue}1}}}{1}
    {2}{{{\color{blue}2}}}{1}
    {3}{{{\color{blue}3}}}{1}
    {4}{{{\color{blue}4}}}{1}
    {5}{{{\color{blue}5}}}{1}
    {6}{{{\color{blue}6}}}{1}
    {7}{{{\color{blue}7}}}{1}
    {8}{{{\color{blue}8}}}{1}
    {9}{{{\color{blue}9}}}{1}
}

\title{When RAG Fails to Equalize: Geo-bias in Factual \\
Question Answering over Public Companies}

\author{
Abhinav Havaldar, Enrico Santus \\
\hspace{.5mm} Bloomberg \\
\hspace{.5mm} \texttt{esantus@bloomberg.net}
}

\begin{document}

\ifcolmsubmission
\linenumbers
\fi

\maketitle
\begin{abstract}
Retrieval-augmented generation (RAG) is widely assumed to mitigate factual errors in large language models (LLMs), but it remains unclear whether retrieval uniformly compensates for missing knowledge.
We study this question in a controlled factual QA setting over public companies, constructing a benchmark of $\sim$2,000 firms across global equity indices. We evaluate six LLMs on four atomic attributes under four conditions: no-context, perfect context, misleading context, and distraction context.
We find strong geographic disparities in no-context accuracy, indicating uneven parametric knowledge. While perfect context improves performance, it does not eliminate these gaps: gains are correlated with baseline accuracy, suggesting retrieval effectiveness is coupled to internal representations. Under misleading context, models frequently copy incorrect information. Larger models improve overall performance but do not remove these structural effects.
These results challenge the view of RAG as a universal corrective and highlight the interaction between model knowledge, context quality, and entity representation.
\end{abstract}

\section{Introduction}

Large language models (LLMs) are increasingly deployed as interfaces for factual question answering in professional workflows. In domains such as finance, corporate intelligence, due diligence, and internal knowledge search, users routinely ask seemingly simple questions—\emph{Where is this company headquartered? When was it founded? Who are its key leaders? What industry does it operate in?} While these queries appear straightforward, they expose two core challenges: parametric knowledge is incomplete and uneven, and retrieval-augmented generation (RAG) does not always behave as a faithful evidence consumption mechanism.

RAG is widely used to improve factuality by supplying external evidence at inference time \citep{lewis2020rag, guu2020realm, borgeaud2022retro}. Although it yields strong average gains, it remains unclear when retrieval genuinely compensates for missing knowledge and when it instead reinforces what models already know well.

\begin{figure}[t]
  \centering
  \begin{minipage}{0.50\textwidth}
    \centering
    \includegraphics[width=\linewidth]{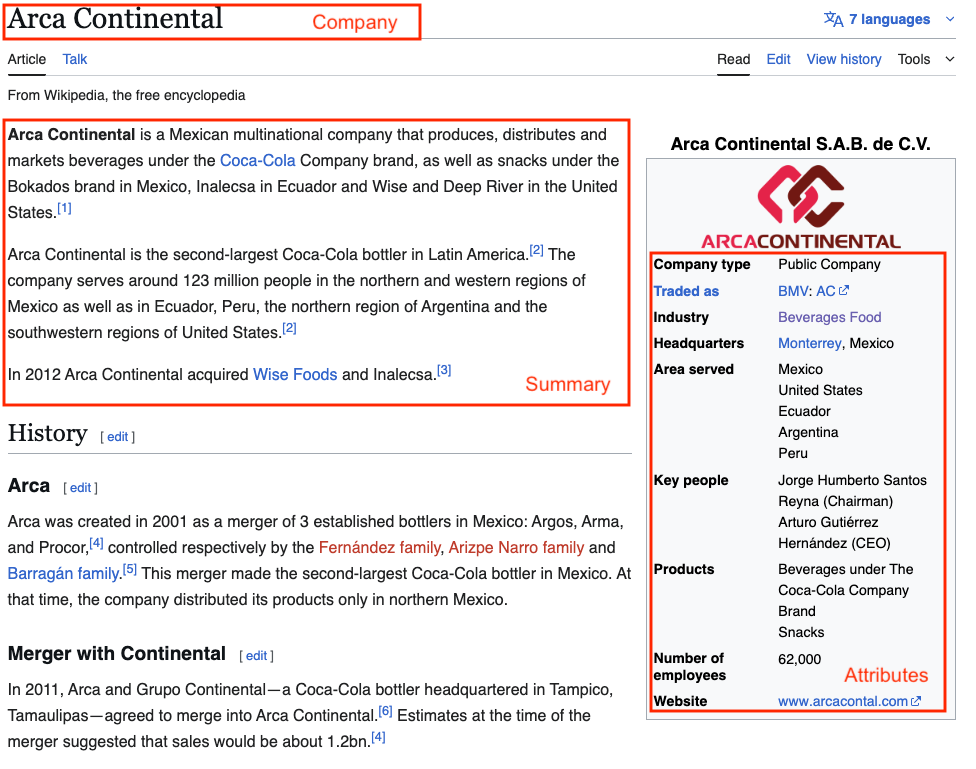}
  \end{minipage}
  \hfill
  \begin{minipage}{0.40\textwidth}
    \begin{lstlisting}[
      language=JSON,
      basicstyle=\ttfamily\scriptsize,
      linewidth=\linewidth,
      breaklines=true,
      frame=single,
      backgroundcolor=\color{gray!6}
    ]
{
 'Company': 'Arca Continental',
 'Industry': 'Drink Industry...',
 'Headquarters': 'Monterrey, Mexico'
}
    \end{lstlisting}
  \end{minipage}

  \vspace{6pt}

  \fbox{%
    \begin{minipage}{0.95\textwidth}
      \scriptsize
      \textbf{Inductive:} Where is Arca Continental headquartered? \\
      \textbf{Deductive:} Which company is headquartered in Monterrey, Mexico?
    \end{minipage}
  }

  \caption{Paired question construction from the same fact.}
  \label{fig:paired-inductive-deductive}
\end{figure}

This question is especially important in domains with uneven factual coverage. Public companies provide a natural example: firms in large, English-dominant markets have rich documentation and strong representation in training data, while companies in smaller or emerging markets often have sparse or fragmented coverage. As a result, identical queries can differ substantially in difficulty depending on the entity involved.

This raises a key issue: retrieval may not act as an independent corrective. If it were, providing perfect evidence should reduce performance gaps across markets. Instead, if retrieval effectiveness depends on existing knowledge—through entity salience or representation quality—it may reinforce existing disparities. In the worst case, imperfect context introduces a second failure mode, where models copy misleading information. Prior work supports this concern, showing a tension between internal priors and external evidence \citep{clasheval2024, parklee2024robustralms}.

At the same time, factual recall is known to be uneven across entities and domains \citep{petroni2019lama, roberts2020cbqa, kandpal2023longtail}. Recent work further highlights geographic disparities in LLM knowledge \citep{moayeri2024worldbench, decoupes2025geodistortions}. However, existing benchmarks largely focus on aggregate accuracy, leaving open how retrieval interacts with these disparities in realistic settings.

We address this gap by introducing a structured evaluation framework for factual QA over public companies. Using Wikipedia, we construct a benchmark of $\sim$2,000 companies across 15 global equity indices and evaluate six LLMs on four corporate attributes (Industry, Founding Year, Headquarters, Key People). As shown in Figure~\ref{fig:paired-inductive-deductive}, each fact is queried in paired inductive and deductive form. To isolate the interaction between parametric knowledge and evidence, we evaluate four conditions: \emph{no-context}, \emph{perfect context}, \emph{misleading context}, and \emph{distraction context}. The benchmark dataset is available at \href{https://zenodo.org/records/19359640}{https://zenodo.org/records/19359640}.

Our results reveal five key findings. First, no-context accuracy varies substantially across markets, reflecting uneven parametric knowledge. Second, inductive questions are consistently easier than deductive ones. Third, perfect context improves performance but does not eliminate disparities, with gains correlated to baseline knowledge. Fourth, misleading context induces systematic copying errors. Fifth, model scale improves overall performance but does not remove these structural effects.

These findings suggest that retrieval is not a universal corrective. Instead, factual reliability depends on the interaction between model knowledge, context quality, and entity representation. This has direct implications for both evaluation and deployment: beyond average accuracy, systems must be assessed for robustness across markets and sensitivity to imperfect evidence.

Our contributions are fourfold:
\begin{enumerate}[leftmargin=*]
    \item A benchmark for factual QA over public companies across 15 global equity indices.
    \item A controlled framework separating parametric knowledge from contextual effects.
    \item Evidence that retrieval gains are coupled with baseline knowledge rather than being uniform.
    \item Identification of systematic failure under misleading context.
\end{enumerate}

\section{Related Work}

A large body of work shows that pretrained language models encode substantial factual knowledge in their parameters. Early probing studies demonstrated that masked language models can recover many relational facts without retrieval \citep{petroni2019lama}, while closed-book QA work framed model parameters as a form of compressed factual memory that improves with scale \citep{roberts2020cbqa}. However, this knowledge is uneven: factual recall is brittle under paraphrase, sensitive to prompt variation, and degrades on rare or long-tail entities \citep{elazar2021pararel, kandpal2023longtail}. Popularity bias in factual recall and entity-centric QA predates the LLM era \citep{pope2021entities, mallen2023popqa, lazaridou2022popularitybias}. These limitations are directly relevant to our setting, as public companies vary widely in salience and textual footprint, making them a natural testbed for structured long-tail knowledge across geography and markets.

Retrieval-augmented generation (RAG) was introduced to address the limits of parametric memory by conditioning models on external documents at inference time \citep{lewis2020rag}. Subsequent work integrated retrieval into pretraining and large-scale modeling, while dense retrieval systems and reader architectures established strong baselines for knowledge-intensive QA \citep{guu2020realm, karpukhin2020dpr, izacard2021fid, borgeaud2022retro}. Benchmark suites such as KILT unified evaluation across tasks \citep{petroni2021kilt}, complementing datasets like Natural Questions and FEVER \citep{kwiatkowski2019natural, thorne2018fever}. While this literature demonstrates that retrieval improves average performance, it typically treats retrieval as an exogenous benefit. Our work instead focuses on \emph{differential gain}: whether retrieval helps uniformly across markets or remains conditioned by existing knowledge disparities.

Recent work questions whether models use retrieved context faithfully. Systems may hallucinate, ignore, or over-copy evidence \citep{gao2023rarr, min2023factscore, huang2023hallucination}. Controlled evaluations show a \emph{tug-of-war} between internal priors and external context: models may follow incorrect evidence when prior knowledge is weak \citep{clasheval2024}, and even strong models can be misled by conflicting context \citep{clasheval2024}. More broadly, imperfect retrieval and adversarial evidence degrade reliability in retrieval-augmented systems \citep{parklee2024robustralms}. Our framework builds on this line of work but grounds it in a real-world domain and leverages geographic heterogeneity to study how evidence use depends on representational inequality.

Bias and uneven representation in NLP are well documented across social and linguistic dimensions \citep{blodgett2020power, hovy2021fivesources}. More recent work shows that LLM factual recall varies systematically across countries and income groups \citep{moayeri2024worldbench}, with measurable geographic distortions in model knowledge \citep{decoupes2025geodistortions}. We extend this perspective from country-level recall to entity-level factual QA. Public companies provide a structured domain where geography, market prominence, and information availability intersect.

Factual reliability is critical in finance, where LLMs are increasingly used for research and decision support. Existing benchmarks such as FinQA and TAT-QA focus on numerical reasoning over financial reports \citep{chen2021finqa, zhu2021tatqa}, while surveys highlight ongoing challenges in factuality and robustness \citep{lee2024finllm}. Our work addresses a complementary problem: atomic factual QA over companies, where the primary challenge lies in uneven coverage and sensitivity to context across global markets.

\section{Research Questions and Hypotheses}

\begin{figure}[t]
\centering
\begin{tikzpicture}[
  font=\footnotesize,
  node distance=7mm and 10mm,
  box/.style={draw, rounded corners=2pt, thick, align=center, inner sep=3pt},
  link/.style={circle, draw, thick, inner sep=0pt, minimum size=6mm},
  arrow/.style={-Latex, line width=0.7pt},
  mod/.style={-Latex, dashed, line width=0.7pt}
]
\node[box] (X) {Geo/Index\\\textbf{$X$}};
\node[box, above=of X] (P) {Model\\Scale\\\textbf{$P$}};
\node[box, right=of X] (K) {Parametric\\Knowledge\\\textbf{$K$}};
\node[link, right=of K] (Q) {\textbf{$Q$}};
\node[box, right=of Q] (Y) {Answer\\Accuracy\\\textbf{$Y$}};
\node[box, below=of Q] (S) {\textbf{Prompt settings}\\
Context \textbf{$C$}: none / perfect / misleading / distraction\\
Direction: inductive / deductive};

\draw[arrow] (X) -- node[below] {H2} (K);
\draw[arrow] (P) -- node[right] {H5} (K);
\draw[arrow] (K) -- (Q) -- (Y);
\draw[arrow] (X) |- node[pos=0.25, left] {$X \rightarrow C$} (S.west);
\draw[mod] (S) -- node[right] {H1/H3/H4} (Q);
\end{tikzpicture}
\vspace{-4pt}
\caption{Hypothesis model. Geo/index position and model scale shape baseline parametric knowledge. Prompt settings determine the instantiated question through context regime and direction, thereby moderating how knowledge and evidence translate into answer accuracy.}
\label{fig:hypothesis-model}
\end{figure}
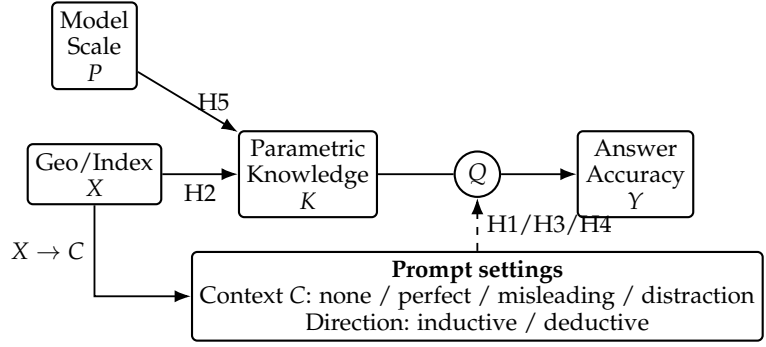

We study how retrieval interacts with unevenly distributed knowledge by treating global market variation as a natural experiment. We analyze factual QA across indices, question formulations, and context regimes to disentangle parametric knowledge, evidence utilization, and robustness to imperfect context. Our goal is to determine whether retrieval acts as an independent corrective mechanism or remains conditioned by structural biases in model knowledge. The hypotheses are illustrated in Figure~\ref{fig:hypothesis-model} and listed below.

\begin{enumerate}
  \item[\textbf{H1}] \textbf{Directional asymmetry:} How does question direction (entity $\rightarrow$ attribute vs.\ attribute $\rightarrow$ entity) affect difficulty and error patterns?
  \item[\textbf{H2}] \textbf{Geo-economic disparity without context:} How does factual accuracy in the absence of retrieval context vary across indices and regions?
  \item[\textbf{H3}] \textbf{Non-uniform benefit of retrieval:} Does perfect context improve accuracy uniformly, or are gains correlated with baseline performance?
  \item[\textbf{H4}] \textbf{Contextual over-reliance:} To what extent does misleading or distraction context induce systematic errors or copying behavior?
  \item[\textbf{H5}] \textbf{Scale and evidence use:} Does model scale improve the use of helpful context and robustness to misleading context, particularly in underrepresented markets?
\end{enumerate}

\section{Methodology}

\subsection{Benchmark Construction}

\paragraph{Data sources and entity collection.}
We construct a benchmark of 2,135 unique publicly listed companies sourced from Wikipedia and mapped to 15 major global equity indices spanning North America, Europe, Asia, Latin America, Africa, and Oceania. For each index, we collect constituent companies and associate each entity with its corresponding index, country, and region, enabling geographically stratified analysis.

The resulting benchmark is intentionally heterogeneous in both market size and information coverage, ranging from large English-dominant markets (e.g., S\&P 500, ASX 200) to mixed-resource settings (e.g., DAX, TSX, Hang Seng) and smaller or emerging markets (e.g., Ghana, Chile, Mexico, Brazil). A full index-level breakdown is provided in Table~\ref{tab:index_combined}.

The dataset contains 2,165 total records due to multi-index membership, corresponding to 2,135 unique companies.

\paragraph{Attribute extraction and normalization.}
For each company, we extract four atomic factual attributes from Wikipedia infobox fields: \textit{headquarters}, \textit{founding year}, \textit{industry}, and \textit{key people}. These attributes are selected for their relevance in corporate analysis and broad availability across entities.

\begin{table}[t]
\centering
\small
\begin{tabular}{lcccc}
\toprule
Variable & Count & Unique & Freq & Top \\
\midrule
\multicolumn{5}{c}{\textit{Dataset Metadata}} \\
\midrule
Company & 2165 & 2135 & 2   & -- \\
Index   & 2165 & 15   & 498 & S\&P 500 (US) \\
\midrule
Industry      & 1703 & 643  & 147 & Financial Serv. \\
Founding Year & 1896 & 212  & 45  & 1994 \\
Headquarters  & 1902 & 666  & 160 & Tokyo, Japan \\
Key People & 1619 & 1594 & 2 & Peter Konieczny \\
\bottomrule
\end{tabular}
\caption{Summary statistics of the dataset. The top panel reports dataset-level metadata, while the bottom panel summarizes the target attributes used for question construction. Note that companies can be listed multiple times in different indices and therefore there are fewer (2135) unique companies than the total (2165). }
\label{tab:attribute_stats}
\end{table}

Table~\ref{tab:attribute_stats} summarizes both dataset metadata and the distribution of target attributes. The high cardinality of attributes such as \textit{industry} and \textit{key people}, combined with partial missingness, highlights the open-ended and heterogeneous nature of the benchmark.

To ensure consistency, we normalize all fields: headquarters are converted to a canonical city-country format; founding year is resolved to a single four-digit value; industry labels are standardized using linked page titles and mapped to a controlled taxonomy; and list-valued attributes (e.g., key people) are whitespace-normalized, title-cased, and sorted into canonical order. This normalization is critical to avoid spurious mismatches (e.g., ``Paris'' vs.\ ``Paris, France'') that would otherwise introduce evaluation noise. Detailed preprocessing steps are described in Appendix~\ref{app:preprocessing}.

\paragraph{Question construction and distractors.}
As shown in Figure~\ref{fig:paired-inductive-deductive}, each fact is converted into two multiple-choice questions that differ only in the direction of inference: \textbf{Inductive} (entity $\rightarrow$ attribute) and \textbf{Deductive} (attribute $\rightarrow$ entity). This paired design enables controlled analysis of directional asymmetry.

Each question is presented with one correct answer and three distractors. Distractors are selected to be semantically plausible but factually incorrect: company summaries are embedded using \texttt{bge-large-en-v1.5}, and for each target we retrieve nearby companies with different attribute values. Depending on the question direction, answer options are instantiated as company names or attribute values. This construction preserves semantic difficulty and reduces the likelihood of solving the task via shallow lexical cues.

Across all companies and attributes, this procedure yields approximately 15{,}000--17{,}000 multiple-choice questions, depending on attribute availability.

\paragraph{Context construction and context regimes.}
For each company, we use the lead paragraph of its Wikipedia page as contextual input. This provides a concise, standardized natural-language summary that contains relevant facts and remains fully reproducible.

To disentangle parametric knowledge from contextual evidence, each question is evaluated under four conditions:
\begin{enumerate}
    \item \textbf{No-context}: model answers using parametric knowledge only.
    \item \textbf{Perfect context}: the correct company summary is provided.
    \item \textbf{Misleading context}: the masked Wikipedia summary of a distractor 
    company is substituted, with \texttt{[COMPANY]} replaced by the target company 
    name, creating locally coherent but factually incorrect evidence 
    (see Appendix~\ref{app:context}).
    \item \textbf{Distraction context}: the summary includes additional irrelevant 
    information.
\end{enumerate}

These regimes allow us to isolate how models balance internal knowledge with external evidence, and to evaluate their robustness to misleading or irrelevant context.

\subsection{Experiment Design}

\paragraph{Models and inference setup.}
We evaluate six language models spanning proprietary and open-weight families: GPT-5, GPT-5 mini, GPT-5 nano, Claude Sonnet 4, LLaMA-70B, and LLaMA-8B. All models are queried using a fixed multiple-choice prompt format with consistent decoding settings. Responses are mapped to answer options and evaluated for correctness. Full model details and inference parameters are provided in Appendix~\ref{app:models}.

We evaluate four context conditions: \textbf{no-context}, \textbf{perfect context}, \textbf{misleading context}, and \textbf{distraction context}, which respectively test parametric recall, evidence utilization, and robustness to incorrect or irrelevant information.

\paragraph{Evaluation metrics.}
We evaluate model responses as binary correctness variables $x_i \in \{0,1\}$ and compute accuracy $\hat{p} = \frac{1}{n}\sum_{i=1}^{n} x_i$ with standard error $\sqrt{\hat{p}(1-\hat{p})/n}$. Confidence intervals are reported to reflect estimation uncertainty.

\paragraph{Stratified analysis.}
To study how context interacts with prior knowledge, we compute baseline no-context accuracy for each index $m$, denoted $\hat{p}_m$, and use it as a proxy for parametric knowledge. We then group indices into bins based on $\hat{p}_m$ and compute context-specific accuracy within each bin:
\[
\hat{p}_{\text{context}, b} =
\frac{\sum_{i=1}^{n} x_{\text{context}, i} \cdot \mathbb{I}(\hat{p}_{m_i} \in (b_{\min}, b_{\max}])}
{\sum_{i=1}^{n} \mathbb{I}(\hat{p}_{m_i} \in (b_{\min}, b_{\max}])}.
\]

\paragraph{Context sensitivity.}
We measure how predictions change relative to the no-context baseline using:
\begin{itemize}
  \item \textbf{Correction rate:} incorrect $\rightarrow$ correct with context.
  \item \textbf{Misleading rate:} correct $\rightarrow$ incorrect under false context.
  \item \textbf{Distraction rate:} correct $\rightarrow$ incorrect under irrelevant context.
\end{itemize}

\paragraph{Statistical modeling.}
We model correctness for each example $i$ and model $m$ as
\[
Y_i^{(m)} \sim \mathrm{Bernoulli}(p),
\]
and compare probabilities across conditions to test five hypotheses:
\begin{itemize}
    \item[H1)] Directional asymmetry: $p$ differs between inductive and deductive questions.
    \item[H2)] Geographic variation: $p$ varies across indices $g$.
    \item[H3)] Dependence on prior knowledge: accuracy with perfect context varies with baseline accuracy.
    \item[H4)] Contextual errors: misleading and distraction effects depend on baseline knowledge.
    \item[H5)] Model effects: $p$ varies across models $m$.
\end{itemize}



\section{Results}

\paragraph{Directional asymmetry.}
We analyze whether question direction affects difficulty. Table~\ref{tab:inductive-vs-deductive} shows that inductive questions are consistently easier than deductive ones across models. This asymmetry reflects two factors: inductive questions resemble direct factual lookup, while deductive questions require identifying the correct entity among plausible alternatives, making them more sensitive to distractors and entity coverage. Deductive reasoning is therefore a stricter test of discriminative knowledge. The gap is not uniform across models, with smaller models showing the largest asymmetry and larger models narrowing it.

We next analyze no-context performance to characterize baseline parametric knowledge across markets.

\begin{table}[t]
\centering
\small
\begin{tabular}{lrrrr}
\toprule
\makecell{Model} & \makecell{Deductive\\Accuracy} & \makecell{Inductive\\Accuracy} & \makecell{Odds Ratio\\(95\% CI)} & \makecell{$p$-value} \\
\midrule
LLaMA-8B     & 0.57 & 0.67 & 0.66 [0.62, 0.70] & 0.00 \\
LLaMA-70B    & 0.70 & 0.76 & 0.75 [0.71, 0.81] & 0.00 \\
Claude Sonnet 4 & 0.78 & 0.80 & 0.76 [0.71, 0.82] & 0.00 \\
GPT-5 nano   & 0.82 & 0.82 & 0.92 [0.86, 0.99] & 0.02 \\
GPT-5 mini   & 0.89 & 0.89 & 1.00 [0.93, 1.08] & 1.00 \\
GPT-5$^\dagger$   & 0.73 & 0.78 & 1.01 [0.92, 1.10] & 0.87 \\
\bottomrule
\end{tabular}
\caption{Directional effect of question framing. Deductive questions are harder than inductive ones for most models. $^\dagger$GPT-5's lower deductive accuracy relative to GPT-5 mini is an exception to this trend; we did not identify a parsing or evaluation artifact underlying it and report it as observed.}
\label{tab:inductive-vs-deductive}
\end{table}

\begin{figure*}[t]
    \centering
    \includegraphics[width=\linewidth]{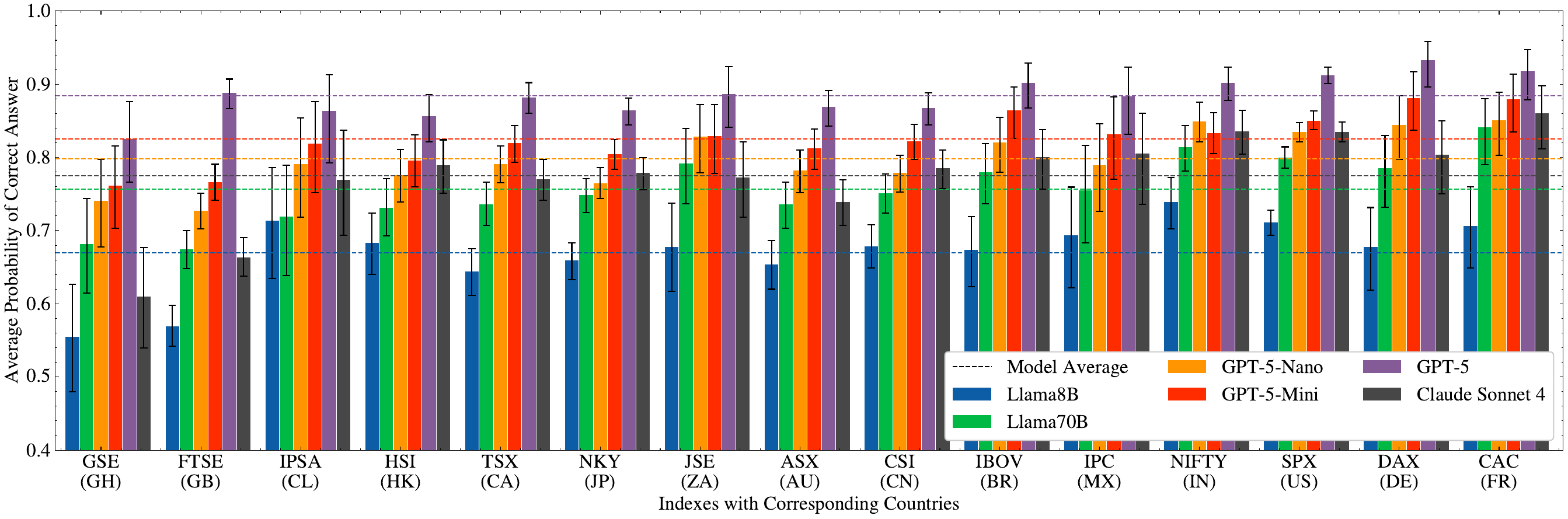}
    \caption{No-context factual accuracy across global equity indices. Grouped bars show the expected probability of a correct answer, with uncertainty intervals, for each model within each index. Horizontal reference lines indicate each model's global mean.}
    \label{fig:results-unconditional}
\end{figure*}

\begin{figure}[t] 
    \centering
    \begin{subfigure}[b]{0.49\textwidth}
        \centering
        \includegraphics[width=\textwidth]{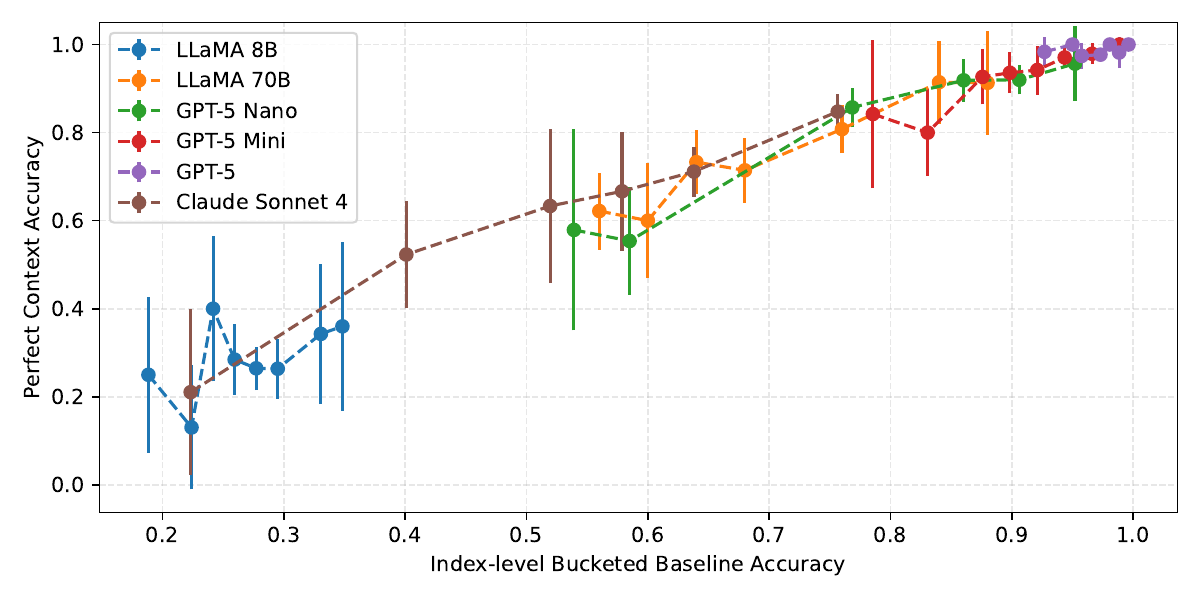}
        \caption{Founding Year (Appendix Table \ref{tab:founding-year-perf-cont-nums})}
    \end{subfigure}
    \begin{subfigure}[b]{0.49\textwidth}
        \centering
        \includegraphics[width=\textwidth]{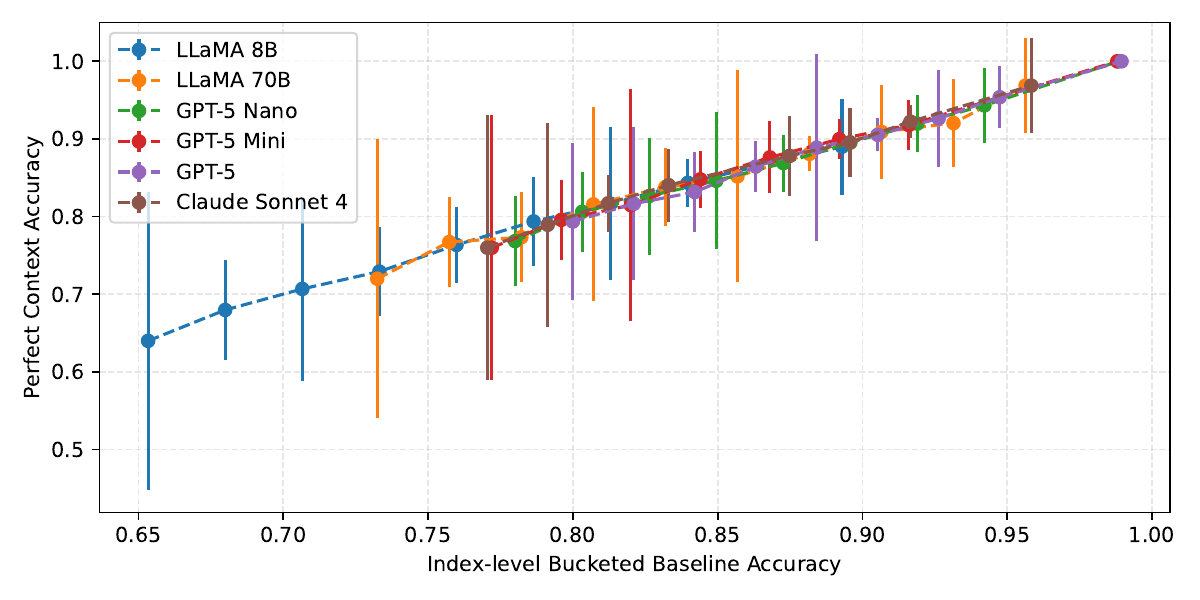}
        \caption{Industry (Appendix Table \ref{tab:sector-perf-cont-nums})}
    \end{subfigure}
    
    \begin{subfigure}[b]{0.49\textwidth}
        \centering
        \includegraphics[width=\textwidth]{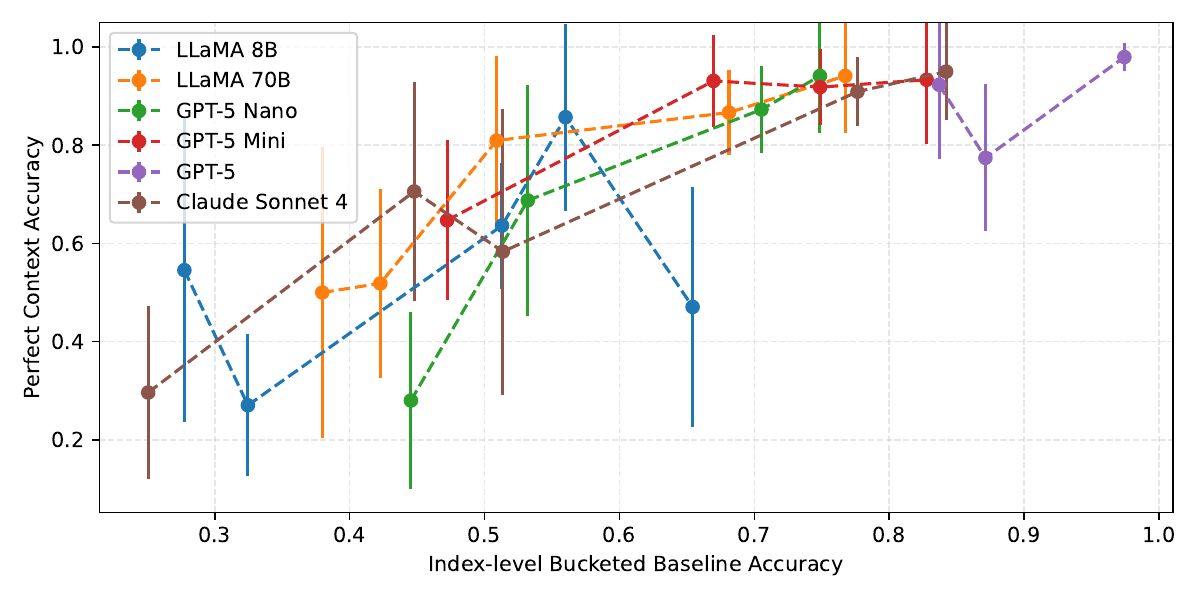}
        \caption{Key People (Appendix Table \ref{tab:people-perf-cont-nums})}
    \end{subfigure}
    \begin{subfigure}[b]{0.49\textwidth}
        \centering
        \includegraphics[width=\textwidth]{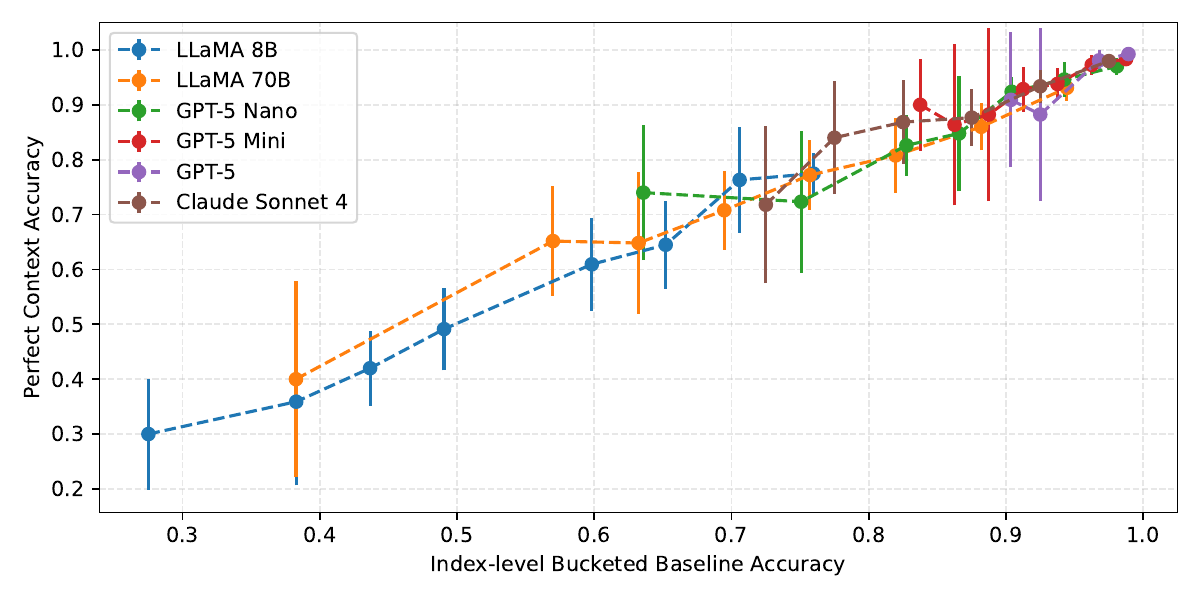}
        \caption{Headquarters (Appendix Table \ref{tab:hq-perf-cont-nums})} 
    \end{subfigure}
    \caption{Effect of perfect context across indices. Positive shifts indicate that models benefit from provided evidence, with different magnitude by markets and model families.}
    \label{fig:results-rag}
\end{figure}



\paragraph{No-context performance varies sharply across markets.}
Figure~\ref{fig:results-unconditional} reports no-context factual accuracy by index and reveals pronounced cross-market variation. For every model, performance differs substantially across indices, with higher accuracy in large, English-dominant markets and lower accuracy in smaller or less-represented ones. This indicates that parametric knowledge is systematically shaped by representational exposure, rather than being uniformly distributed. The consistency of this pattern across both proprietary and open-weight models suggests that it reflects a general property of LLMs rather than model-specific behavior. While larger models achieve higher overall accuracy, they preserve the same relative disparities, indicating that scale improves performance but does not eliminate uneven knowledge distribution. Attribute-level no-context results are provided in Appendix Figure~\ref{fig:gdp-accuracy}.



\paragraph{Perfect context improves performance, but not uniformly.}
Figure~\ref{fig:results-rag} shows that supplying correct contextual evidence improves accuracy across all models, confirming that retrieval can enhance factual QA. However, these gains are not uniform: markets with stronger no-context performance tend to remain stronger even with context. If retrieval acted as an independent corrective, accuracy would converge across markets; instead, we observe a coupling between baseline parametric knowledge and evidence utilization. This pattern is consistent across attributes and models, with a modest dampening effect for larger models. The corresponding stratified numerical results are reported in Appendix Tables~\ref{tab:founding-year-perf-cont-nums}--\ref{tab:hq-perf-cont-nums}.



\begin{figure}[htbp]
    \centering
    \begin{subfigure}[b]{0.49\textwidth}
        \centering
        \includegraphics[width=\textwidth]{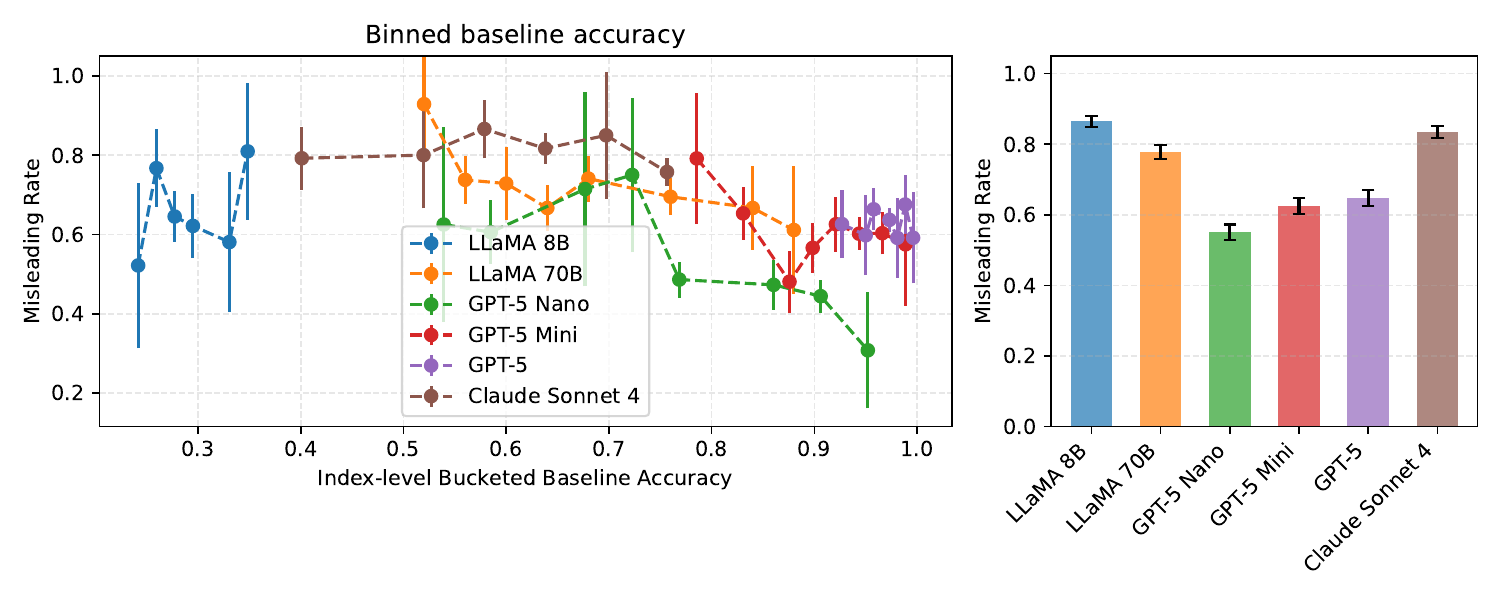}
        \caption{Founding Year (Appendix Table \ref{tab:founding-year-misl-cont-nums})}
    \end{subfigure}
    \begin{subfigure}[b]{0.49\textwidth}
        \centering
        \includegraphics[width=\textwidth]{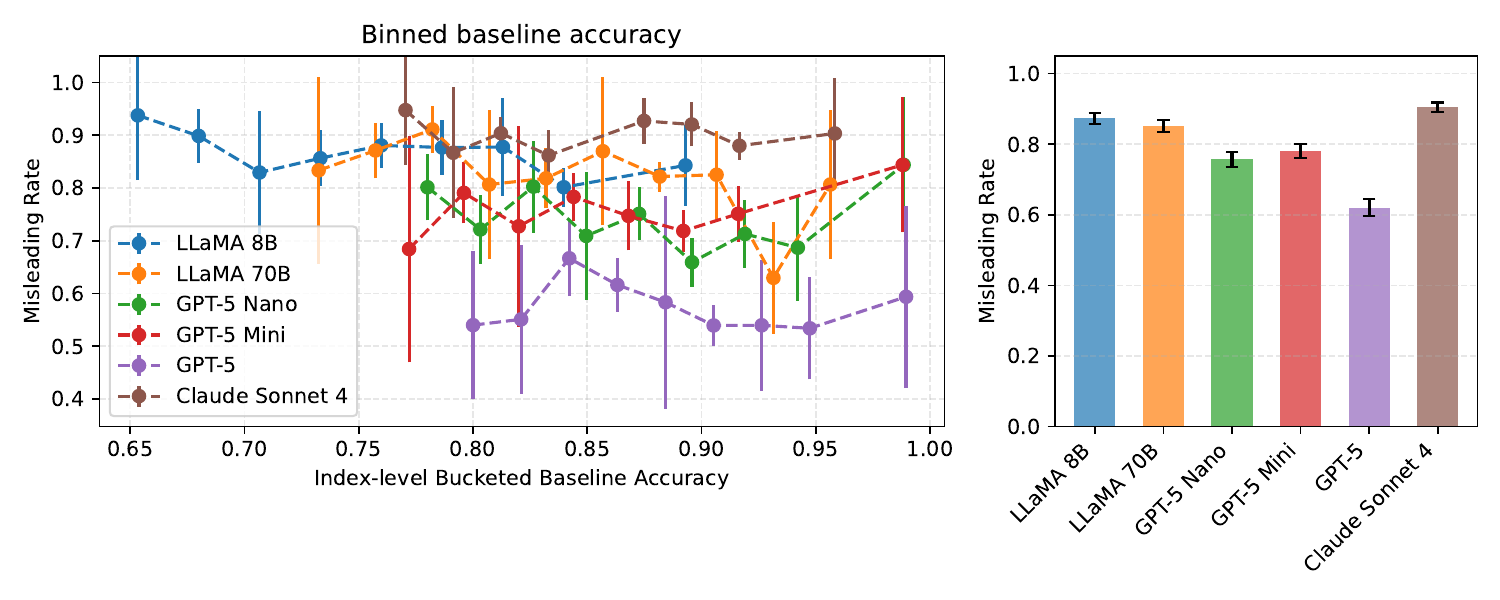}
        \caption{Industry (Appendix Table \ref{tab:sector-misl-cont-nums})}
    \end{subfigure}
        
    \begin{subfigure}[b]{0.49\textwidth}
        \centering
        \includegraphics[width=\textwidth]{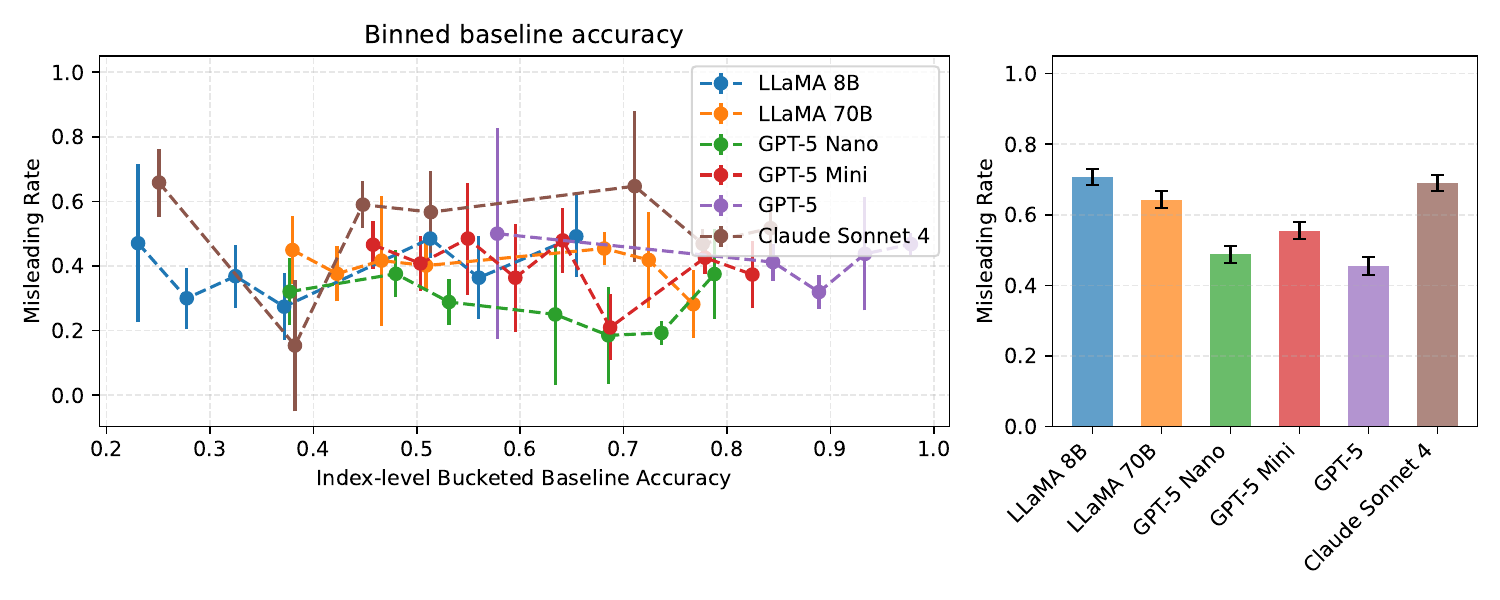}
        \caption{Key People (Appendix Table \ref{tab:people-misl-cont-nums})}
    \end{subfigure}
    \begin{subfigure}[b]{0.49\textwidth}
        \centering
        \includegraphics[width=\textwidth]{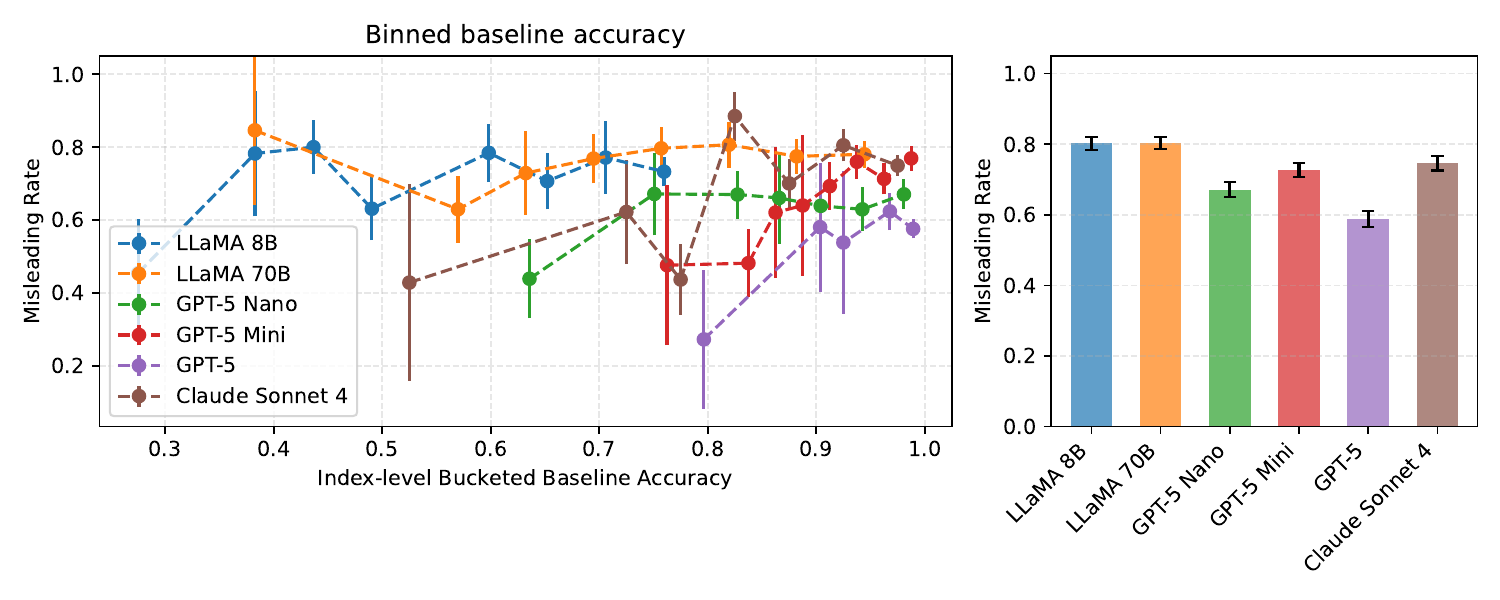}
        \caption{Headquarters (Appendix Table \ref{tab:hq-misl-cont-nums})}
    \end{subfigure}
    \caption{Effect of misleading context across indices. Decreasing misleading rates indicate the ability of models to use prior information to avoid relying on false information.}
    \label{fig:misleading-results}
\end{figure}

\paragraph{Misleading context induces systematic copying failures.}
Figure~\ref{fig:misleading-results} shows that when context is misleading, accuracy drops substantially, as models often adopt incorrect evidence rather than resist it. This is not merely the absence of benefit but an active degradation: models can perform worse with incorrect context than under parametric recall alone. This pattern aligns with prior-context conflict findings in RAG systems, here observed in a globally heterogeneous domain. Models that follow context too readily may appear grounded while remaining brittle, which is particularly problematic in high-stakes settings where visible evidence can create false trust. While stronger parametric knowledge can sometimes mitigate this effect, the pattern is inconsistent and depends on the task. In general, larger models show greater robustness, though they remain susceptible to misleading information. The corresponding stratified numerical results are reported in Appendix Tables~\ref{tab:founding-year-misl-cont-nums}--\ref{tab:hq-misl-cont-nums}.


\begin{figure}[htbp]
    \centering
    \begin{subfigure}[b]{0.49\textwidth}
        \centering
        \includegraphics[width=\textwidth]{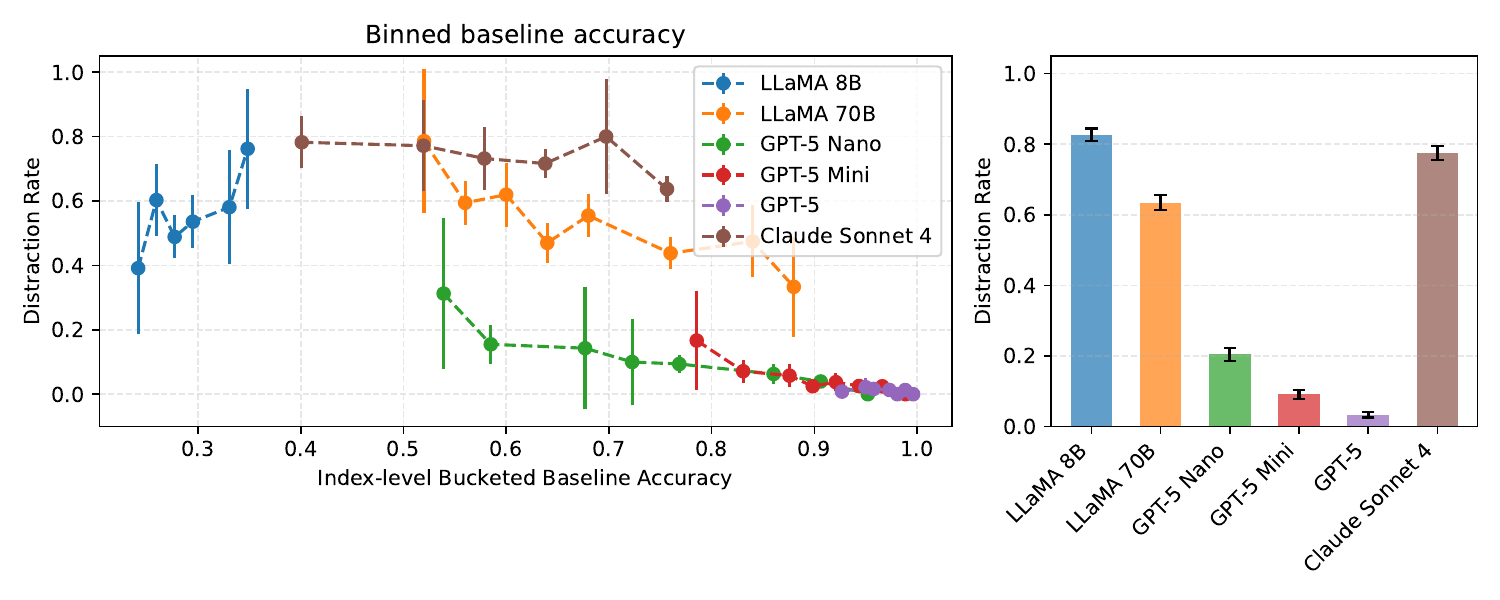}
        \caption{Founding Year (Appendix Table \ref{tab:founding-year-dist-cont-nums})}
    \end{subfigure}
    \begin{subfigure}[b]{0.49\textwidth}
        \centering
        \includegraphics[width=\textwidth]{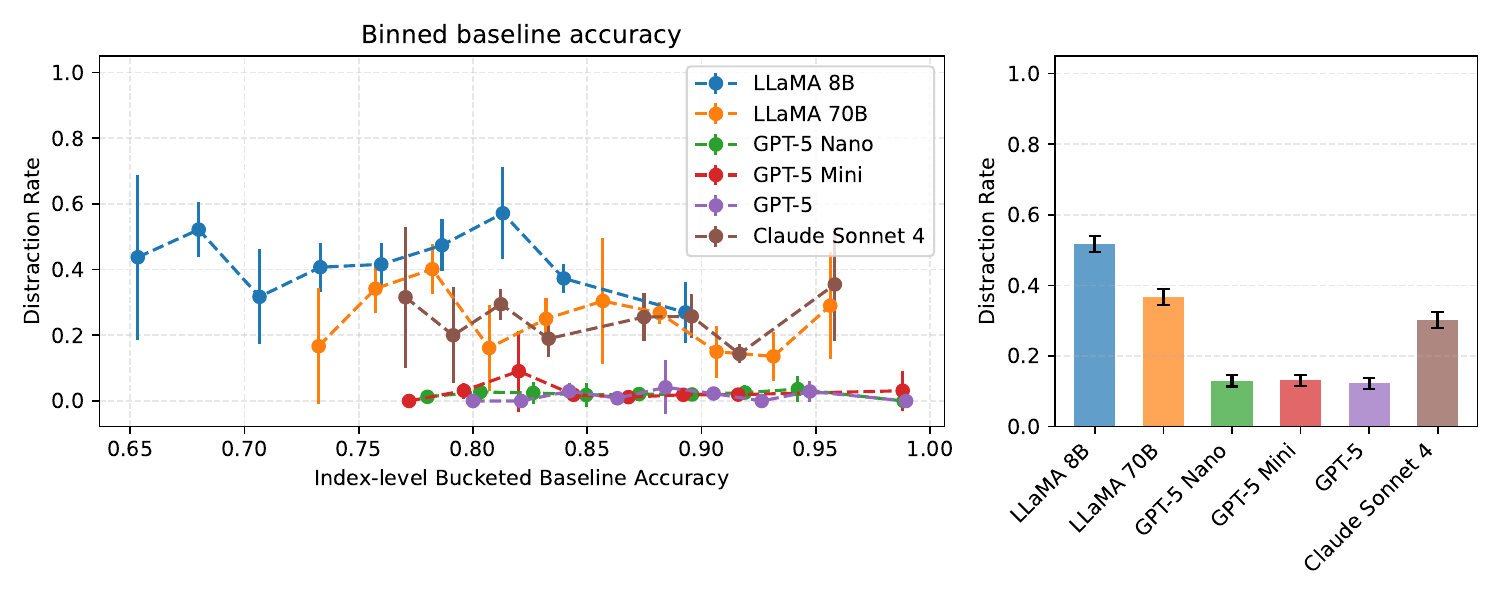}
        \caption{Industry (Appendix Table \ref{tab:sector-dist-cont-nums})}
    \end{subfigure}    
    \begin{subfigure}[b]{0.49\textwidth}
        \centering
        \includegraphics[width=\textwidth]{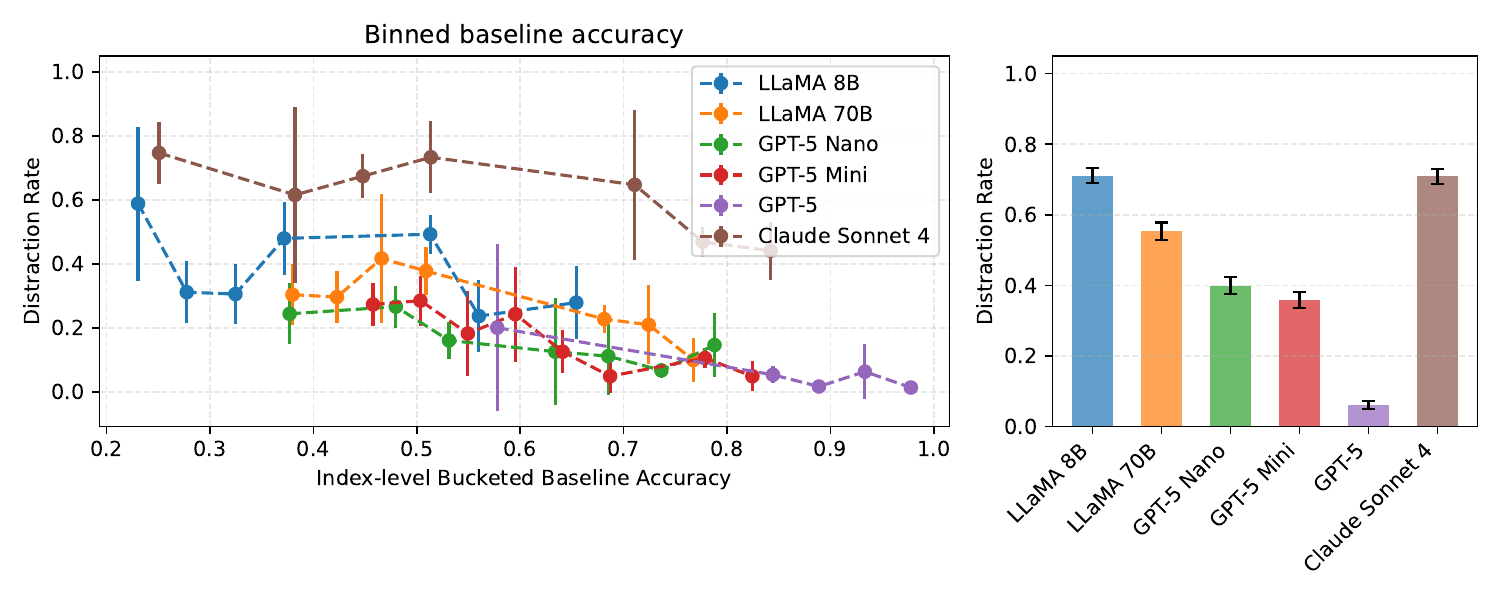}
        \caption{Key People (Appendix Table \ref{tab:people-dist-cont-nums})}
    \end{subfigure}
    \begin{subfigure}[b]{0.49\textwidth}
        \centering
        \includegraphics[width=\textwidth]{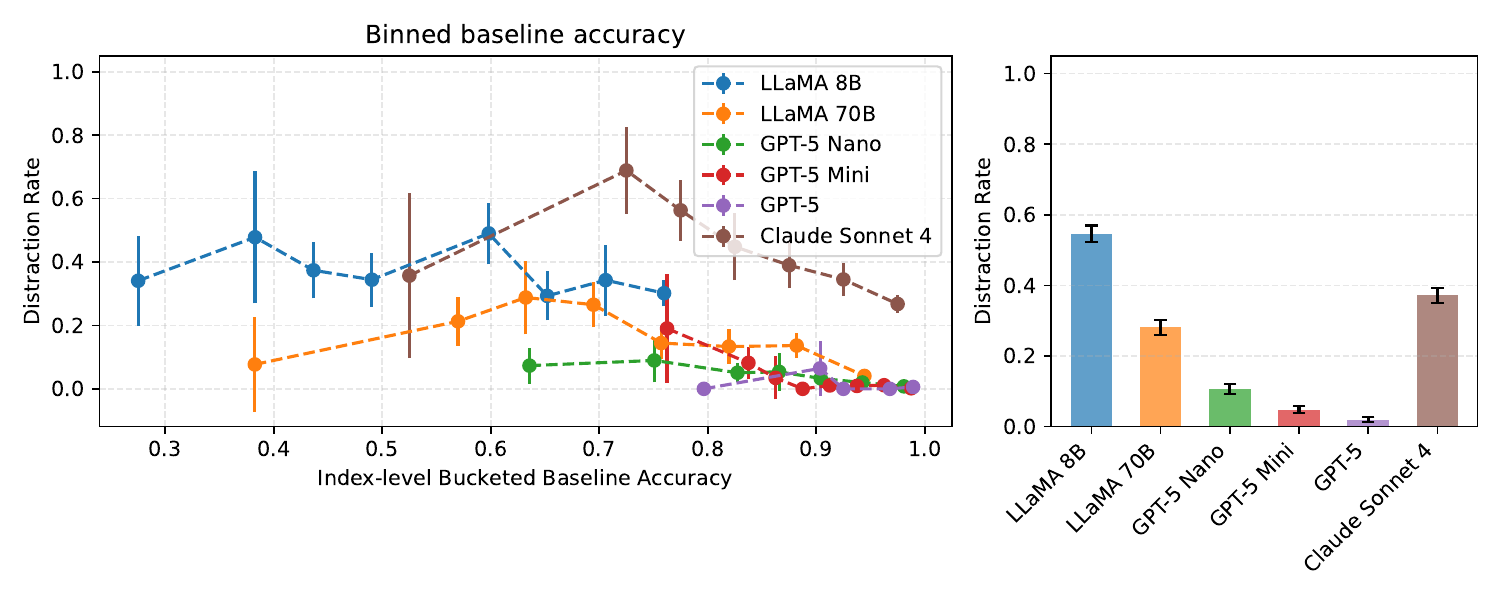}
        \caption{Headquarters (Appendix Table \ref{tab:hq-dist-cont-nums})}
    \end{subfigure}
    \caption{Effect of distracting context across indices. Decreasing distraction rates indicate the ability of models to use prior information to avoid relying on irrelevant information.}
    \label{fig:distraction-results}
\end{figure}

\paragraph{Distraction context degrades performance more mildly but systematically.}
Figure~\ref{fig:distraction-results} shows a similar pattern under irrelevant rather 
than false context. The effect is more consistent and generally weaker than in the 
misleading setting, suggesting that models are more vulnerable to locally plausible 
false evidence than to merely irrelevant text. Larger models and indices with stronger 
baseline knowledge are more robust overall, though distractor difficulty may be 
attenuated in non-English-dominant markets due to the English-centric embedding model 
used for distractor selection. The corresponding stratified numerical results are 
reported in Appendix Tables~\ref{tab:founding-year-dist-cont-nums}--\ref{tab:hq-dist-cont-nums}.

\paragraph{Model scale improves performance but does not remove inequality.}
Across all conditions, larger models outperform smaller ones in absolute terms: they achieve higher baseline accuracy, benefit more from helpful context, and are more robust to misleading or distracting evidence. This suggests that scale strengthens both factual recall and evidence utilization.

However, scale does not alter the qualitative structure of the results. The same markets remain comparatively easy or difficult, and patterns of retrieval coupling and contextual brittleness persist. Thus, scale acts as an additive improvement—shifting performance upward without eliminating underlying geo-economic disparities in knowledge and evidence use.

\section{Discussion}

Taken together, the results support all five hypotheses: factual QA performance reflects an interaction between parametric knowledge and contextual evidence. \textbf{H1}: inductive questions are easier than deductive ones. \textbf{H2}: no-context accuracy varies sharply across indices, reflecting uneven knowledge. \textbf{H3}: correct context improves accuracy but does not close these gaps, with gains tied to baseline knowledge. \textbf{H4}: misleading context degrades performance, showing over-reliance on incorrect evidence. \textbf{H5}: larger models improve performance and robustness without removing structural disparities.

These results challenge retrieval as a universal corrective. Instead, it is a conditional amplifier: helping where models are already strong, while offering less benefit and more fragility where knowledge is sparse. Because aggregate accuracy hides systematic weaknesses, evaluation should be stratified by geography, entity coverage, and context condition; robustness to misleading evidence should be a core requirement, motivating validation layers, structured data integration, and provenance-aware design in high-stakes domains.



\section{Conclusion}
We introduced a benchmark for factual QA over public companies that uses geo-economic heterogeneity to study knowledge and retrieval in large language models. Accuracy is uneven across markets; correct context narrows but does not eliminate these gaps because gains remain tied to baseline accuracy; and misleading context induces systematic errors, a failure mode that persists even as larger models improve.

Overall, retrieval is not a universal corrective, underscoring the need for stratified, robustness-oriented evaluation in high-stakes settings. The benchmark dataset is available at \href{https://zenodo.org/records/19359640}{https://zenodo.org/records/19359640}.

\paragraph{Limitations.} The benchmark relies on Wikipedia-derived fields and summaries---reproducible, but unlike production retrieval---and targets atomic facts rather than long-form or multi-hop reasoning. The misleading-context condition is synthetic; real-world errors may be subtler. Coverage quality and distractor difficulty may vary across markets, a confound multilingual or locally sourced corpora could reduce. Finally, the benchmark is English-centric and limited to US/European models; extending to non-Western models, open-ended factuality evaluation, and an ``I don't know'' option for calibration are natural next steps.

\section*{Acknowledgments}
  We would like to thank Sebastian Gehrmann, Ivailo Dimov, David Rosenberg, and Arun Verma for the insightful discussions, which helped improve this paper.

\bibliography{references}
\bibliographystyle{colm2026_conference}

\appendix

\section{Data}

\subsection{Dataset Retrieval and Composition}
\label{app:index_breakdown}
\label{app:data_retrieval}

Data retrieval proceeds in two stages. First, we collect Wikipedia pages corresponding to the target equity indices and parse their constituent company lists. Second, for each company page, we extract the lead summary paragraph and relevant infobox fields. Duplicate entities are merged after normalization.

Table~\ref{tab:index_combined} summarizes the dataset composition and corresponding Wikipedia sources.

\FloatBarrier

\begin{table}[t]
\centering
\setlength{\tabcolsep}{3pt}
\begin{tabular}{llllr}
\toprule
Index & Country & Reg & Wiki Page & \# \\
\midrule
GSE & Ghana & AF & \texttt{GSE\_Composite\_Index} & 33 \\
N225 & Japan & AS & \texttt{Nikkei\_225} & 458 \\
FTSE & United Kingdom & EU & \texttt{FTSE\_250\_Index} & 250 \\
IBOV & Brazil & LA & \texttt{List\_of\_companies\_listed\_on\_B3} & 68 \\
ASX200 & Australia & OC & \texttt{S\&P/ASX\_200} & 146 \\
HSI & Hong Kong & AS & \texttt{Hang\_Seng\_Index} & 81 \\
JSE & South Africa & AF & \texttt{FTSE/JSE\_Top\_40\_Index} & 37 \\
TSX & Canada & NA & \texttt{S\&P/TSX\_Composite\_Index} & 171 \\
IPSA & Chile & LA & \texttt{Indice\_de\_Precios\_Selectivo\_de\_Acciones} & 23 \\
CSI300 & China & AS & \texttt{CSI\_300\_Index} & 191 \\
IPC & Mexico & LA & \texttt{Indice\_de\_Precios\_y\_Cotizaciones} & 30 \\
SPX & United States & NA & \texttt{List\_of\_S\&P\_500\_companies} & 498 \\
DAX & Germany & EU & \texttt{DAX} & 41 \\
NIFTY50 & India & AS & \texttt{NIFTY\_50} & 98 \\
FRA & France & EU & \texttt{CAC\_40} & 40 \\
\midrule
Total & -- & -- & -- & 2165 \\
\bottomrule
\end{tabular}
\caption{Dataset composition and sources (Reg: region).}
\label{tab:index_combined}
\end{table}

\subsection{Company pages}
For each company page, we extract the lead paragraph and the fields used to instantiate the four target attributes as shown in Figure \ref{fig:paired-inductive-deductive}. Each of the constituent companies was accessed on 2026-01-03. We omit all but the four fields (Headquarters, Founding Year, Key People, and Industry) along with the company's name and leading paragraph.

\subsection{Preprocessing and Normalization}
\label{app:preprocessing}
Many of the fields are not normalized (i.e., headquarters may be “Paris” or “Paris, France”), which can cause confusion. To avoid these issues, we more aggressively normalize our raw data. In a few cases, to parse the strings, we use \texttt{LLaMA 3.1 70B} with $temperature = 0.02$. We select infobox fields that are both common enough and semantically appropriate for atomic factual QA. Fields with extreme sparsity or highly unstable formatting are excluded.

\begin{table}[t]
    \centering
    \small
    \begin{tabular}{lrr}
        \toprule
         & Missingness & Uniqueness \\
        \midrule
        \rowcolor{gray!25} Company Name & 0.00 & 96.98 \\
        \rowcolor{gray!25} Index Name & 0.00 & 0.81 \\
        \rowcolor{gray!25} Summary & 0.54 & 94.74 \\
        Website & 8.95 & 90.17 \\
        \rowcolor{gray!25} Headquarters & 12.19 & 68.55 \\
        \rowcolor{gray!25} Founding Year & 13.70 & 72.75 \\
        Company Type & 14.94 & 5.33 \\
        Traded As & 15.91 & 93.97 \\
        \rowcolor{gray!25} Industry & 16.40 & 43.87 \\
        \rowcolor{gray!25} Key People & 16.94 & 94.09 \\
        Revenue & 25.62 & 93.33 \\
        Number of Employees & 26.91 & 91.37 \\
        Net Income & 29.61 & 92.80 \\
        Total Assets & 45.36 & 93.29 \\
        \rowcolor{gray!25} Products & 46.55 & 83.75 \\
        Operating Income & 48.81 & 92.62 \\
        Total Equity & 50.81 & 92.43 \\
        \rowcolor{gray!25} Area Served & 60.84 & 25.48 \\
        Subsidiaries & 68.45 & 93.33 \\
        Founder & 71.41 & 91.51 \\
        \bottomrule
    \end{tabular}
    \caption{Raw Wikipedia-derived fields, with fields used in this study highlighted in gray.}
    \label{tab:attributes}
\end{table}

\subsubsection{Normalizing Founding Year}
We resolve the founding field to the first valid four-digit year after removing markup and auxiliary text.

\subsubsection{Normalizing Key People}
These list-valued fields are first split by their delimiters, then stripped of excess whitespace, title-cased, and finally sorted into a canonical order.

\subsubsection{Normalizing Industry}
Industry is by far the most open-ended field in this study and required multiple normalization steps. 
Industry items are normalized using linked page titles where available. All values are then title-cased and sorted. The prompt used for determining the GICS sector is as follows:

\begin{quote}
    Look at the unstructured industries mentioned in the following text and extract a list of all the industries for this company. Resolve each to one of the Global Industry Classification Standard (GICS) sectors:

- Communication Services

- Consumer Discretionary

- Consumer Staples

- Energy

- Financials

- Health Care

- Industrials

- Information Technology

- Materials

- Real Estate

- Utilities

Respond with a JSON object such as \{``industries'': [``Health Care'', ``Financials'']\}
\end{quote}

The results are then title-cased, and duplicates are removed. We also verify if the results are in the list of GICS sectors. Finally, we sort the industries lexically to give a canonical list.

\subsubsection{Normalizing Headquarters}
Each location string is split on commas and reduced to city and country by passing it to the aforementioned model with the following prompt.
\begin{quote}
    Extract the city and country (Full Name) as a list of JSON objects. Leave any unknown field blank. \\

Response should follow \{ ``locations'': [\{``city'': ``Sydney'', ``country'': ``Australia''\}]\}
\end{quote}

Country aliases are mapped to canonical names by a human reviewer who resolves names like ``People's Republic of China'' to ``China''. Furthermore, we resolve all variations caused by differences in diacritics and accents. 

\subsubsection{Normalizing Summaries}
The leading and trailing whitespace is removed from each leading paragraph on the Wikipedia pages. We then query the aforementioned LLM to identify all direct mentions of the company and mask them with \texttt{[COMPANY]} using the following prompt:

 \begin{quote}
Consider the following summary for \{\{company\}\} and mask all mentions of \{\{company\}\} with [COMPANY]. Do not change any other part of the text.\\
Summary: \{\{summary\}\}

Response should follow \{ ``masked\_summary'': ``...'' \}
     
 \end{quote}

Each of the responses is then compared to the original to ensure that all non-mention text remains unchanged. This additional field is referred to as the \texttt{masked\_summary}.

\section{Methodology}
\subsection{Prompt Templates}
\label{app:prompts}

Below is the prompt structure used for multiple-choice evaluation. 

\subsubsection{No-context Prompt}
\begin{quote}
You are answering a factual multiple-choice question about a public company.

Question: \{\{question\}\}

A) \{\{choice\_A\}\}

B) \{\{choice\_B\}\}

C) \{\{choice\_C\}\}

D) \{\{choice\_D\}\}

Select the most appropriate choice and respond with only one of A, B, C, or D as a JSON object like \{ ``response'': ``A'' \}.
\end{quote}

\subsubsection{Perfect/Misleading/Distraction Context Prompt}
\begin{quote}
You are answering a factual multiple-choice question about a public company. Use the provided context if helpful.

Context: \{\{context\}\}

Question: \{\{question\}\}

A) \{\{choice\_A\}\}

B) \{\{choice\_B\}\}

C) \{\{choice\_C\}\}

D) \{\{choice\_D\}\}

Select the most appropriate choice and respond with only one of A, B, C, or D as a JSON object like \{ ``response'': ``A'' \}.
\end{quote}

\subsubsection{Context Construction}
\label{app:context}

Correct-context examples use the company's original Wikipedia lead paragraph. Misleading-context examples are generated by replacing the target attribute value in the lead paragraph with an incorrect but plausible value drawn from another entity, while preserving the surrounding text as much as possible. This creates minimally perturbed evidence that is locally coherent but factually wrong with respect to the queried attribute.

\subsection{Constructing Contexts}
To study the effects of providing contexts with varying veracity and relevance, we construct perfect, misleading, and distraction contexts. These contexts are constructed per field, per company, and per index, and correspond to the same question asked with different additional information. This section details how these contexts are constructed.

\subsubsection{Perfect Context}
The perfect context was constructed using the unchanged leading paragraph from each company’s Wikipedia page. For certain attributes, the summary did not contain the fact being tested. For example, the founding year may not be mentioned in the summary for Apple. To control for these absences, we filter our questions and responses for the perfect context scenario. These filters are specific to the fields being queried.

\begin{itemize}
    \item Headquarters: We use the regex \texttt{ \textbackslash b\{\{city\}\}\textbackslash b} to verify if the city of the correct headquarters is mentioned in the summary. 
    \item Founding Year: We use the regex \texttt{ \textbackslash b\{\{founding year\}\}\textbackslash b} to verify if the founding year of the company is mentioned in the summary.
    \item Key People: We use the regex \texttt{ \textbackslash b\{\{last name\}\}\textbackslash b} to verify if the last name of the correct key person is mentioned in the summary.
    \item Industry: Due to the central nature of the industry of a company, we make the assumption that each summary contains at least some information regarding the industry. Hence, no filtering is applied to questions regarding Industry. 
    \item Across the benchmark, these filters exclude 61.0\% of Founding Year questions, 29.6\% of Headquarters questions, and 86.0\% of Key People questions from the perfect-context condition, reflecting how rarely Wikipedia lead paragraphs restate infobox facts verbatim (Table~\ref{tab:exclusion-by-index}). Exclusion is highest for Key People across every index, consistent with lead paragraphs rarely naming executives by surname.
\end{itemize}
    
\begin{table}[h!]
    \centering
    \small
    \begin{tabular}{lccc}
    \toprule
    Index & Founding Year & Headquarters & Key People \\
    \midrule
    GSE & 38.7 & 73.3 & 100.0 \\
    N225 & 56.6 & 31.0 & 93.2 \\
    FTSE & 73.0 & 29.7 & 84.9 \\
    IBOV & 54.5 & 31.8 & 87.9 \\
    ASX200 & 65.2 & 62.6 & 94.0 \\
    HSI & 60.5 & 31.2 & 79.7 \\
    JSE & 55.9 & 32.4 & 97.1 \\
    TSX & 64.3 & 25.4 & 85.5 \\
    IPSA & 76.2 & 56.5 & 95.0 \\
    CSI300 & 62.0 & 29.5 & 96.3 \\
    IPC & 78.6 & 39.3 & 76.0 \\
    SPX & 58.7 & 18.7 & 82.7 \\
    DAX & 41.5 & 17.1 & 80.0 \\
    NIFTY50 & 63.2 & 21.6 & 70.3 \\
    FRA & 37.5 & 25.6 & 72.5 \\
    \bottomrule
    \end{tabular}
    \caption{Perfect-context exclusion rate (\%) by attribute and index.}
    \label{tab:exclusion-by-index}
\end{table}

\subsubsection{Misleading Context}
To construct a purposefully misleading context, we follow these steps:
\begin{enumerate}
    \item Select any of the incorrect options for this question.
    \item Find its \texttt{masked\_summary}.
    \item Replace all mentions of \texttt{[COMPANY]} with the name of the company we are studying.
\end{enumerate}
Consider the example in which we ask the model about Apple’s industry and have provided the industry for Johnson \& Johnson (Health Care) as a choice. A misleading context here would replace all mentions of Johnson \& Johnson in its summary with Apple, as shown below.
\begin{quote}
    \textbf{Apple} is an American multinational pharmaceutical, biotechnology, and medical technologies corporation headquartered in New Brunswick, New Jersey, and publicly traded on the New York Stock Exchange. Its common stock is a component of the Dow Jones Industrial Average, and the company is ranked No. 48 on the 2025 Fortune 500 list of the largest United States corporations. In 2025, the company was ranked 42nd in the Forbes Global 2000. \textbf{Apple} has a global workforce of approximately 138,000...
\end{quote}

\subsubsection{Distraction Context}
The distraction context is constructed in the same manner as the misleading context with one small change. Instead of replacing the name of the target company, the incorrect company’s name is left in the context. Using the same example as above, we leave Johnson \& Johnson in the summary untouched.
\subsection{Robustness and Coverage Diagnostics}
\label{app:diagnostics}

\paragraph{Summary length.} Mean Wikipedia lead-paragraph length varies by index (60--158 words) but does not significantly predict perfect-context accuracy (Pearson $r = 0.432$, $p = 0.11$, $n=15$ indices), suggesting that source-text length is not the primary driver of cross-market performance gaps.

\paragraph{Distractor similarity.} Since distractors are selected using an English-centric embedding model (\texttt{bge-large-en-v1.5}), we test whether this makes distractors easier to rule out for non-English-market companies. Mean TF-IDF cosine similarity between target and distractor summaries is higher for English-market indices (GSE, FTSE, JSE, TSX, ASX200, SPX; mean $=0.061$) than non-English-market indices (mean $=0.046$), though the difference is not significant ($t$-test, $p=0.24$, $n=15$). This does not support a strong confound: distractors are, if anything, no easier for non-English markets.

\paragraph{Company-size (revenue quintile) robustness.} To test whether geographic disparities persist after controlling for company size, we regress answer accuracy on within-index Wikipedia revenue quintiles (quintile 1 = smallest), together with index fixed effects. Across all four context conditions, size is a significant predictor of accuracy (quintile coefficient $p < 0.01$ in every condition; Table~\ref{tab:quintile-regression}), but index fixed effects remain jointly significant after controlling for it ($p < 0.001$ in every condition), and only 1 of 15 indices (SPX) shows strictly monotonic accuracy across quintiles. Geo-economic disparities are therefore not fully explained by company size.

\begin{table}[h!]
\centering
\small
\begin{tabular}{lccc}
\toprule
Condition & Quintile coef. & Quintile $p$ & Index FE $p$ \\
\midrule
No-Context          & 0.036 & $<0.001$ & $<0.001$ \\
Perfect Context     & 0.041 & $<0.001$ & $<0.001$ \\
Distraction Context & 0.040 & $<0.001$ & $<0.001$ \\
Misleading Context  & 0.030 & 0.002    & $<0.001$ \\
\bottomrule
\end{tabular}
\caption{Company-size (revenue quintile) regression, controlling for index fixed effects.}
\label{tab:quintile-regression}
\end{table}

\subsection{Models and Inference Metadata}
\label{app:models}
\subsubsection{Model Details}
Here we specify the hyperparameter settings used when querying the different models, along with the dates on which the models were called and their specific versions. These details are listed in Table \ref{tab:model-specs}.

\FloatBarrier

\begin{table}[t]
    \centering
    \begin{tabular}{lrrrr}
        \toprule
         Model & Parameters & Version & Temperature & Max Output Tokens \\
         \midrule
         LLaMA & 8B & 3.1 & 1.0 & 4,096 \\
         LLaMA & 70B & 3.1 & 1.0 & 8,192 \\
         GPT-5 nano & $-^{*}$ & 5.0 & $-^{+}$ & 128,000 \\
         GPT-5 mini & $-^{*}$ & 5.0 & $-^{+}$ & 128,000 \\
         GPT-5& $-^{*}$ & 5.0 & $-^{+}$ & 128,000 \\
         Claude Sonnet & $-^{*}$ & 4.0 & 1.0 & 64,000 \\
         \bottomrule
    \end{tabular}
    \caption{Language Model Access Details. $^{*}$: Closed source models do not disclose details regarding parameters. $^{+}$: OpenAI does not permit users to specify temperature.}
    \label{tab:model-specs}
\end{table}

Each of these models was accessed using the Python OpenAI Chat Completions API with internal Bloomberg-hosted endpoints. Each question was asked individually in its own query (i.e., the queries were not batched).

\subsubsection{Parsing LLM Responses}
All of our prompts ask the models to respond using JSON objects. We find the first instance of the \texttt{\{} character and the last instance of the \texttt{\}} character. The string enclosed by these two characters is extracted, parsed as a JSON object, and used for the appropriate analysis. Any deviations that make it impossible to parse the JSON response are excluded from the analysis.

\section{Results}

In all result tables, \textit{Center} denotes the midpoint of the corresponding baseline-accuracy bin used for stratified analysis. Rows correspond to quantiles of baseline no-context accuracy, and tables report either \textit{Accuracy}, \textit{Misleading Rate}, or \textit{Distraction Rate}, depending on the evaluation setting, each with its associated uncertainty.

\subsection{No-Context}
We begin by evaluating model performance without external evidence, isolating parametric knowledge. For stratified analyses, companies are grouped into bins according to baseline no-context accuracy.

\begin{figure}[h!]
    \centering
    
    \includegraphics[width=\textwidth]{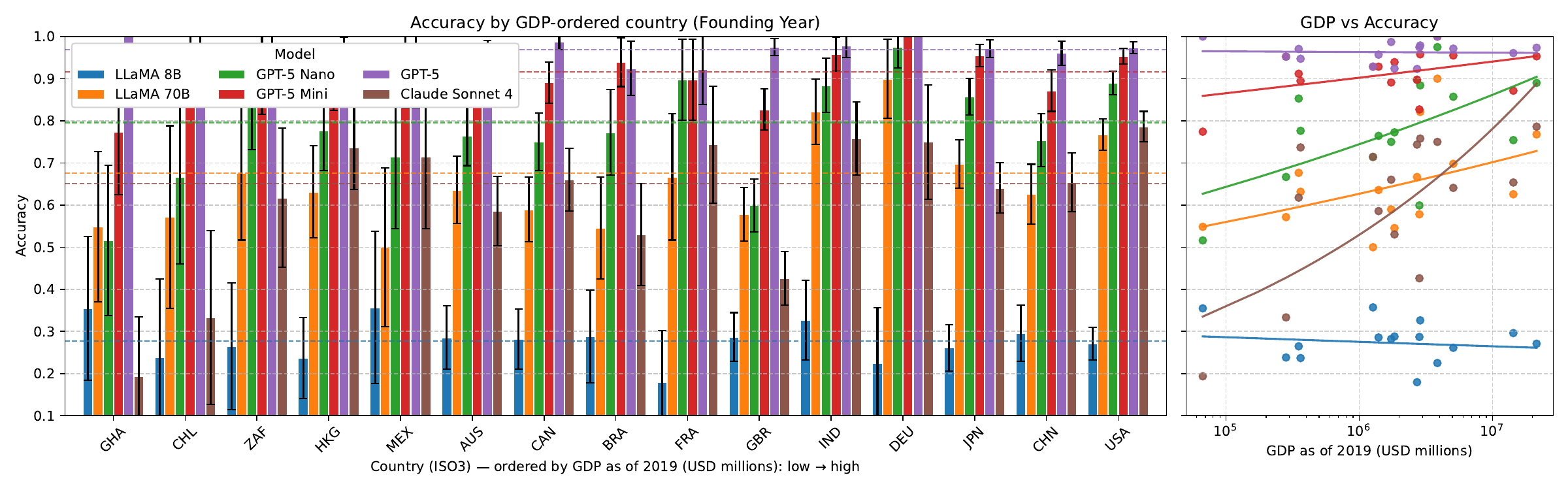}
    \caption*{(a) Founding Year}
    
    \vspace{10pt}
    
    \includegraphics[width=\textwidth]{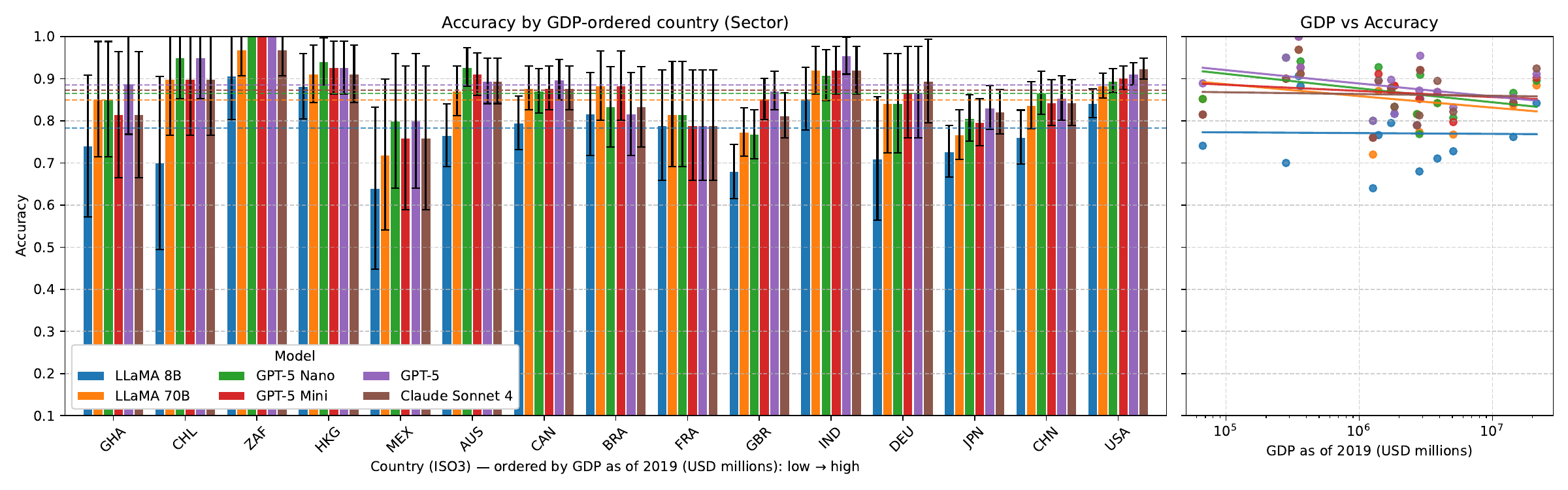}
    \caption*{(b) Industry}
    
    \vspace{10pt}
    
    \includegraphics[width=\textwidth]{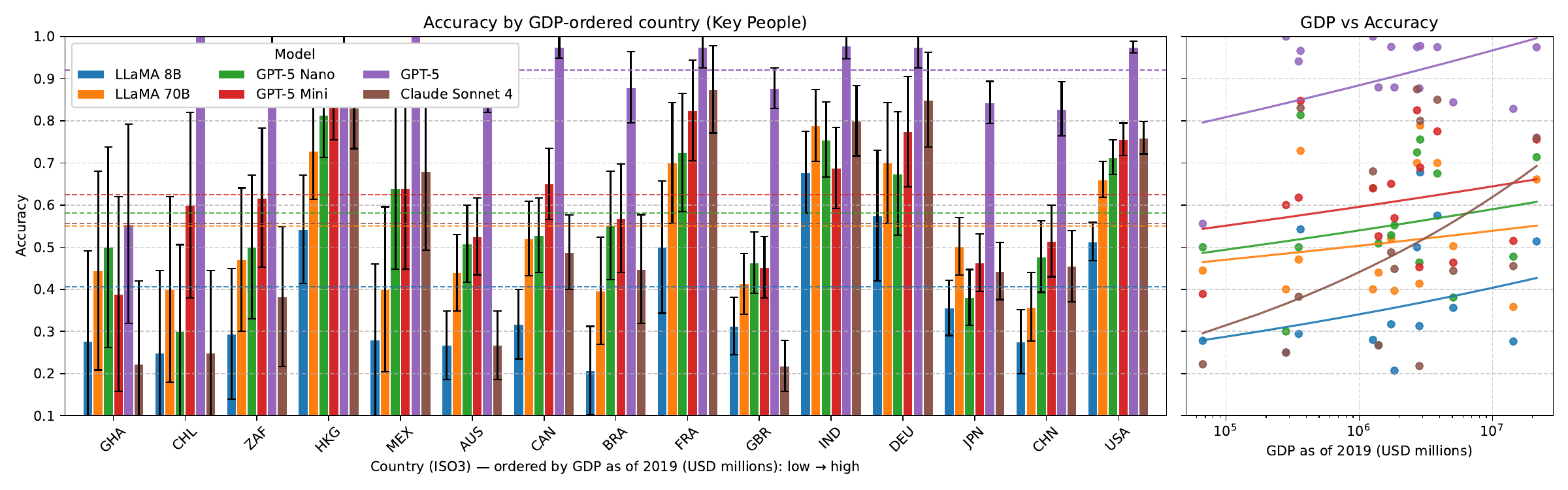}
    \caption*{(c) Key People}
    
    \vspace{10pt}
    
    \includegraphics[width=\textwidth]{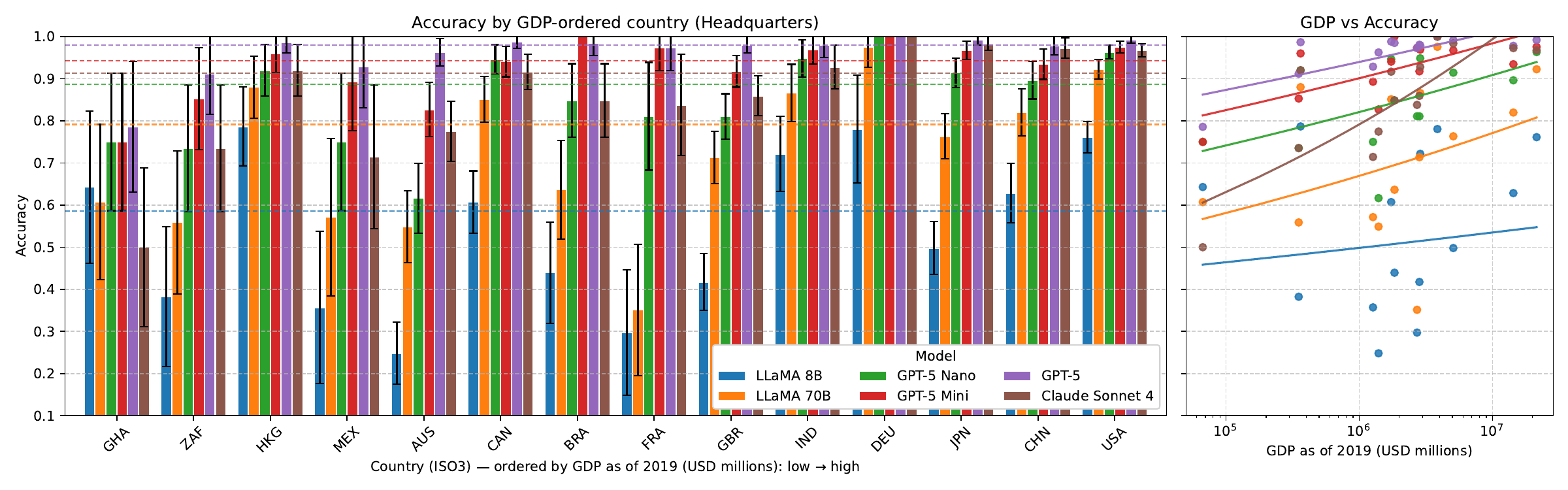}
    \caption*{(d) Headquarters}
    
    \caption{Accuracy by Index by GDP}
    \label{fig:gdp-accuracy}
\end{figure}

Figure~\ref{fig:gdp-accuracy} summarizes no-context performance across attributes and indices.

\FloatBarrier

\FloatBarrier

\subsection{Perfect Context}

We next evaluate performance when correct contextual evidence is provided. This setting measures the extent to which retrieval improves factual QA and whether gains depend on prior knowledge.

Stratified results across attributes are reported in Tables~\ref{tab:founding-year-perf-cont-nums}--\ref{tab:hq-perf-cont-nums}.

\begin{table}[h!]
    \centering
    \resizebox{\textwidth}{!}{%
    \begin{tabular}{lllllllllllll}
    \toprule
    model & \multicolumn{2}{c}{Claude Sonnet 4} & \multicolumn{2}{c}{GPT-5} & \multicolumn{2}{c}{GPT-5 Mini} & \multicolumn{2}{c}{GPT-5 Nano} & \multicolumn{2}{c}{LLaMA-70B} & \multicolumn{2}{c}{LLaMA-8B} \\
     & Accuracy & Center & Accuracy & Center & Accuracy & Center & Accuracy & Center & Accuracy & Center & Accuracy & Center \\
    Quantile &  &  &  &  &  &  &  &  &  &  &  &  \\
    \midrule
    0.1 & 0.21 ± 0.10 & 0.22 & 0.98 ± 0.02 & 0.93 & 0.84 ± 0.09 & 0.79 & 0.58 ± 0.12 & 0.54 & - & - & 0.25 ± 0.09 & 0.19 \\
    0.2 & - & - & - & - & - & - & 0.55 ± 0.06 & 0.58 & 0.62 ± 0.04 & 0.56 & - & - \\
    0.3 & - & - & - & - & 0.80 ± 0.05 & 0.83 & - & - & 0.60 ± 0.07 & 0.60 & 0.13 ± 0.07 & 0.22 \\
    0.4 & 0.52 ± 0.06 & 0.40 & 1.00 ± 0.00 & 0.95 & - & - & - & - & 0.73 ± 0.04 & 0.64 & 0.40 ± 0.08 & 0.24 \\
    0.5 & - & - & 0.97 ± 0.01 & 0.96 & 0.93 ± 0.03 & 0.88 & - & - & 0.71 ± 0.04 & 0.68 & 0.28 ± 0.04 & 0.26 \\
    0.6 & 0.63 ± 0.09 & 0.52 & - & - & 0.94 ± 0.02 & 0.90 & 0.86 ± 0.02 & 0.77 & - & - & 0.26 ± 0.03 & 0.28 \\
    0.7 & 0.67 ± 0.07 & 0.58 & 0.98 ± 0.01 & 0.97 & 0.94 ± 0.03 & 0.92 & - & - & 0.81 ± 0.03 & 0.76 & 0.26 ± 0.03 & 0.29 \\
    0.8 & 0.71 ± 0.03 & 0.64 & 1.00 ± 0.00 & 0.98 & 0.97 ± 0.01 & 0.94 & 0.92 ± 0.02 & 0.86 & - & - & - & - \\
    0.9 & - & - & 0.98 ± 0.02 & 0.99 & 0.98 ± 0.01 & 0.97 & 0.92 ± 0.02 & 0.91 & 0.91 ± 0.05 & 0.84 & 0.34 ± 0.08 & 0.33 \\
    1.0 & 0.85 ± 0.02 & 0.76 & 1.00 ± 0.00 & 1.00 & 1.00 ± 0.00 & 0.99 & 0.96 ± 0.04 & 0.95 & 0.91 ± 0.06 & 0.88 & 0.36 ± 0.10 & 0.35 \\
    \bottomrule
    \end{tabular}
    } 
    \caption{Statistics for Accuracy given Perfect Context for questions about Founding Year.}
    \label{tab:founding-year-perf-cont-nums}
\end{table}
\begin{table}[h!]
    \centering
    \resizebox{\textwidth}{!}{%
    \begin{tabular}{llllllllllrll}
    \toprule
    model & \multicolumn{2}{c}{Claude Sonnet 4} & \multicolumn{2}{c}{GPT-5} & \multicolumn{2}{c}{GPT-5 Mini} & \multicolumn{2}{c}{GPT-5 Nano} & \multicolumn{2}{c}{LLaMA-70B} & \multicolumn{2}{c}{LLaMA-8B} \\
     & Accuracy & Center & Accuracy & Center & Accuracy & Center & Accuracy & Center & Accuracy & Center & Accuracy & Center \\
    Quantile &  &  &  &  &  &  &  &  &  &  &  &  \\
    \midrule
    0.1 & 0.76 ± 0.09 & 0.77 & 0.79 ± 0.05 & 0.80 & 0.76 ± 0.09 & 0.77 & 0.77 ± 0.03 & 0.78 & 0.72 ± 0.09 & 0.73 & 0.64 ± 0.10 & 0.65 \\
    0.2 & 0.79 ± 0.07 & 0.79 & 0.82 ± 0.05 & 0.82 & 0.80 ± 0.03 & 0.80 & 0.81 ± 0.03 & 0.80 & 0.77 ± 0.03 & 0.76 & 0.68 ± 0.03 & 0.68 \\
    0.3 & 0.82 ± 0.02 & 0.81 & 0.83 ± 0.03 & 0.84 & 0.81 ± 0.08 & 0.82 & 0.83 ± 0.04 & 0.83 & 0.77 ± 0.03 & 0.78 & 0.71 ± 0.06 & 0.71 \\
    0.4 & 0.84 ± 0.02 & 0.83 & 0.86 ± 0.02 & 0.86 & 0.85 ± 0.02 & 0.84 & 0.85 ± 0.05 & 0.85 & 0.82 ± 0.06 & 0.81 & 0.73 ± 0.03 & 0.73 \\
    0.5 & - & - & 0.89 ± 0.06 & 0.88 & 0.88 ± 0.02 & 0.87 & 0.87 ± 0.02 & 0.87 & 0.84 ± 0.03 & 0.83 & 0.76 ± 0.02 & 0.76 \\
    0.6 & 0.88 ± 0.03 & 0.87 & 0.91 ± 0.01 & 0.91 & 0.90 ± 0.01 & 0.89 & 0.90 ± 0.01 & 0.90 & 0.85 ± 0.07 & 0.86 & 0.79 ± 0.03 & 0.79 \\
    0.7 & 0.90 ± 0.02 & 0.90 & 0.93 ± 0.03 & 0.93 & 0.92 ± 0.02 & 0.92 & 0.92 ± 0.02 & 0.92 & 0.88 ± 0.01 & 0.88 & 0.82 ± 0.05 & 0.81 \\
    0.8 & 0.92 ± 0.01 & 0.92 & 0.95 ± 0.02 & 0.95 & - & - & 0.94 ± 0.02 & 0.94 & 0.91 ± 0.03 & 0.91 & 0.84 ± 0.02 & 0.84 \\
    0.9 & - & - & - & - & - & - & - & - & 0.92 ± 0.03 & 0.93 & - & - \\
    1.0 & 0.97 ± 0.03 & 0.96 & 1.00 ± 0.00 & 0.99 & 1.00 ± 0.00 & 0.99 & 1.00 ± 0.00 & 0.99 & 0.97 ± 0.03 & 0.96 & 0.89 ± 0.03 & 0.89 \\
    \bottomrule
    \end{tabular}} 
    \caption{Statistics for Accuracy given Perfect Context for questions about Industry.}
    \label{tab:sector-perf-cont-nums}
\end{table}
\begin{table}[h!]
    \centering
    \resizebox{\textwidth}{!}{%
    \begin{tabular}{lllllllllllll}
\toprule
model & \multicolumn{2}{c}{Claude Sonnet 4} & \multicolumn{2}{c}{GPT-5} & \multicolumn{2}{c}{GPT-5 Mini} & \multicolumn{2}{c}{GPT-5 Nano} & \multicolumn{2}{c}{LLaMA-70B} & \multicolumn{2}{c}{LLaMA-8B} \\
 & Accuracy & Center & Accuracy & Center & Accuracy & Center & Accuracy & Center & Accuracy & Center & Accuracy & Center \\
Quantile &  &  &  &  &  &  &  &  &  &  &  &  \\
\midrule
0.1 & 0.30 ± 0.09 & 0.25 & 0.92 ± 0.08 & 0.84 & 0.65 ± 0.08 & 0.47 & - & - & 0.50 ± 0.15 & 0.38 & - & - \\
0.2 & - & - & - & - & - & - & 0.28 ± 0.09 & 0.45 & 0.52 ± 0.10 & 0.42 & 0.55 ± 0.16 & 0.28 \\
0.3 & - & - & 0.77 ± 0.08 & 0.87 & - & - & - & - & - & - & 0.27 ± 0.07 & 0.32 \\
0.4 & 0.71 ± 0.11 & 0.45 & - & - & - & - & 0.69 ± 0.12 & 0.53 & 0.81 ± 0.09 & 0.51 & - & - \\
0.5 & 0.58 ± 0.15 & 0.51 & - & - & - & - & - & - & - & - & - & - \\
0.6 & - & - & - & - & 0.93 ± 0.05 & 0.67 & - & - & - & - & - & - \\
0.7 & - & - & - & - & - & - & - & - & - & - & 0.64 ± 0.07 & 0.51 \\
0.8 & - & - & - & - & 0.92 ± 0.04 & 0.75 & 0.87 ± 0.05 & 0.71 & 0.87 ± 0.04 & 0.68 & 0.86 ± 0.10 & 0.56 \\
0.9 & 0.91 ± 0.04 & 0.78 & 0.98 ± 0.01 & 0.97 & - & - & 0.94 ± 0.06 & 0.75 & - & - & - & - \\
1.0 & 0.95 ± 0.05 & 0.84 & - & - & 0.93 ± 0.07 & 0.83 & - & - & 0.94 ± 0.06 & 0.77 & 0.47 ± 0.12 & 0.65 \\
\bottomrule
\end{tabular}} 
    \caption{Statistics for Accuracy given Perfect Context for questions about Key People.}
    \label{tab:people-perf-cont-nums}
\end{table}

\begin{table}[h!]
    \centering
    \resizebox{\textwidth}{!}{%
    \begin{tabular}{lllllllllllll}
    \toprule
    model & \multicolumn{2}{c}{Claude Sonnet 4} & \multicolumn{2}{c}{GPT-5} & \multicolumn{2}{c}{GPT-5 Mini} & \multicolumn{2}{c}{GPT-5 Nano} & \multicolumn{2}{c}{LLaMA-70B} & \multicolumn{2}{c}{LLaMA-8B} \\
     & Accuracy & Center & Accuracy & Center & Accuracy & Center & Accuracy & Center & Accuracy & Center & Accuracy & Center \\
    Quantile &  &  &  &  &  &  &  &  &  &  &  &  \\
    \midrule
    0.1 & 0.62 ± 0.18 & 0.52 & 0.88 ± 0.12 & 0.80 & 0.75 ± 0.16 & 0.76 & 0.74 ± 0.06 & 0.64 & 0.40 ± 0.09 & 0.38 & 0.30 ± 0.05 & 0.28 \\
    0.3 & - & - & - & - & - & - & - & - & - & - & 0.36 ± 0.08 & 0.38 \\
    0.4 & - & - & - & - & 0.90 ± 0.04 & 0.84 & 0.72 ± 0.07 & 0.75 & 0.65 ± 0.05 & 0.57 & 0.42 ± 0.03 & 0.44 \\
    0.5 & 0.72 ± 0.07 & 0.72 & - & - & 0.86 ± 0.07 & 0.86 & - & - & 0.65 ± 0.07 & 0.63 & 0.49 ± 0.04 & 0.49 \\
    0.6 & 0.84 ± 0.05 & 0.78 & 0.91 ± 0.06 & 0.90 & 0.88 ± 0.08 & 0.89 & 0.83 ± 0.03 & 0.83 & 0.71 ± 0.04 & 0.69 & - & - \\
    0.7 & 0.87 ± 0.04 & 0.82 & 0.88 ± 0.08 & 0.92 & 0.93 ± 0.02 & 0.91 & 0.85 ± 0.05 & 0.87 & 0.77 ± 0.03 & 0.76 & 0.61 ± 0.04 & 0.60 \\
    0.8 & 0.88 ± 0.03 & 0.88 & - & - & 0.94 ± 0.02 & 0.94 & 0.92 ± 0.01 & 0.90 & 0.81 ± 0.03 & 0.82 & 0.64 ± 0.04 & 0.65 \\
    0.9 & 0.93 ± 0.02 & 0.92 & 0.98 ± 0.01 & 0.97 & 0.97 ± 0.01 & 0.96 & 0.95 ± 0.02 & 0.94 & 0.86 ± 0.02 & 0.88 & 0.76 ± 0.05 & 0.71 \\
    1.0 & 0.98 ± 0.01 & 0.98 & 0.99 ± 0.00 & 0.99 & 0.98 ± 0.01 & 0.99 & 0.97 ± 0.01 & 0.98 & 0.93 ± 0.01 & 0.94 & 0.77 ± 0.02 & 0.76 \\
    \bottomrule
    \end{tabular}} 
    \caption{Statistics for Accuracy given Perfect Context for questions about Headquarters.}
    \label{tab:hq-perf-cont-nums}
\end{table}

\subsection{Misleading Context}

We evaluate robustness under misleading context, where provided evidence is locally coherent but factually incorrect.

Tables report the \textit{Misleading Rate}, defined as the proportion of previously correct no-context predictions that become incorrect under misleading context. The \textit{Center} column denotes the midpoint of the corresponding baseline-accuracy bin.

Results are shown in Tables~\ref{tab:founding-year-misl-cont-nums}--\ref{tab:hq-misl-cont-nums}.

\begin{table}
    \centering
    \resizebox{\textwidth}{!}{%
\begin{tabular}{lllllllllllll}
\toprule
model & \multicolumn{2}{r}{Claude Sonnet 4} & \multicolumn{2}{r}{GPT-5} & \multicolumn{2}{r}{GPT-5 Mini} & \multicolumn{2}{r}{GPT-5 Nano} & \multicolumn{2}{r}{LLaMA-70B} & \multicolumn{2}{r}{LLaMA-8B} \\
 & Center & Rate & Center & Rate & Center & Rate & Center & Rate & Center & Rate & Center & Rate \\
Quantile &  &  &  &  &  &  &  &  &  &  &  &  \\
\midrule
0.1 & - & - & 0.93 & 0.63 ± 0.04 & 0.79 & 0.79 ± 0.08 & 0.54 & 0.62 ± 0.12 & 0.52 & 0.93 ± 0.07 & - & - \\
0.2 & - & - & - & - & - & - & 0.58 & 0.61 ± 0.04 & 0.56 & 0.74 ± 0.03 & - & - \\
0.3 & - & - & - & - & 0.83 & 0.65 ± 0.03 & - & - & 0.60 & 0.73 ± 0.05 & - & - \\
0.4 & 0.40 & 0.79 ± 0.04 & 0.95 & 0.60 ± 0.05 & - & - & 0.68 & 0.71 ± 0.13 & 0.64 & 0.67 ± 0.03 & 0.24 & 0.52 ± 0.11 \\
0.5 & - & - & 0.96 & 0.66 ± 0.03 & 0.88 & 0.48 ± 0.04 & 0.72 & 0.75 ± 0.10 & 0.68 & 0.74 ± 0.03 & 0.26 & 0.77 ± 0.05 \\
0.6 & 0.52 & 0.80 ± 0.07 & - & - & 0.90 & 0.57 ± 0.03 & 0.77 & 0.49 ± 0.02 & - & - & 0.28 & 0.65 ± 0.03 \\
0.7 & 0.58 & 0.87 ± 0.04 & 0.97 & 0.64 ± 0.02 & 0.92 & 0.63 ± 0.04 & - & - & 0.76 & 0.69 ± 0.02 & 0.29 & 0.62 ± 0.04 \\
0.8 & 0.64 & 0.82 ± 0.02 & 0.98 & 0.59 ± 0.05 & 0.94 & 0.60 ± 0.02 & 0.86 & 0.47 ± 0.03 & - & - & - & - \\
0.9 & 0.70 & 0.85 ± 0.08 & 0.99 & 0.68 ± 0.04 & 0.97 & 0.60 ± 0.03 & 0.91 & 0.44 ± 0.02 & 0.84 & 0.67 ± 0.05 & 0.33 & 0.58 ± 0.09 \\
1.0 & 0.76 & 0.76 ± 0.02 & 1.00 & 0.59 ± 0.06 & 0.99 & 0.57 ± 0.08 & 0.95 & 0.31 ± 0.07 & 0.88 & 0.61 ± 0.08 & 0.35 & 0.81 ± 0.09 \\
\bottomrule
\end{tabular}
    } 
    \caption{Statistics for Misleading Rate given Misleading Context for questions about Founding Year.}
    \label{tab:founding-year-misl-cont-nums}
\end{table}
\begin{table}
    \centering
    \resizebox{\textwidth}{!}{%
\begin{tabular}{lllllllllllll}
\toprule
model & \multicolumn{2}{r}{Claude Sonnet 4} & \multicolumn{2}{r}{GPT-5} & \multicolumn{2}{r}{GPT-5 Mini} & \multicolumn{2}{r}{GPT-5 Nano} & \multicolumn{2}{r}{LLaMA-70B} & \multicolumn{2}{r}{LLaMA-8B} \\
 & Center & Rate & Center & Rate & Center & Rate & Center & Rate & Center & Rate & Center & Rate \\
Quantile &  &  &  &  &  &  &  &  &  &  &  &  \\
\midrule
0.1 & 0.77 & 0.95 ± 0.05 & 0.80 & 0.54 ± 0.07 & 0.77 & 0.68 ± 0.11 & 0.78 & 0.80 ± 0.03 & 0.73 & 0.83 ± 0.09 & 0.65 & 0.94 ± 0.06 \\
0.2 & 0.79 & 0.87 ± 0.06 & 0.82 & 0.55 ± 0.07 & 0.80 & 0.79 ± 0.03 & 0.80 & 0.72 ± 0.03 & 0.76 & 0.87 ± 0.03 & 0.68 & 0.90 ± 0.03 \\
0.3 & 0.81 & 0.90 ± 0.02 & 0.84 & 0.67 ± 0.04 & 0.82 & 0.73 ± 0.10 & 0.83 & 0.80 ± 0.04 & 0.78 & 0.91 ± 0.02 & 0.71 & 0.83 ± 0.06 \\
0.4 & 0.83 & 0.86 ± 0.02 & 0.86 & 0.62 ± 0.03 & 0.84 & 0.78 ± 0.02 & 0.85 & 0.71 ± 0.06 & 0.81 & 0.81 ± 0.07 & 0.73 & 0.86 ± 0.03 \\
0.5 & - & - & 0.88 & 0.58 ± 0.10 & 0.87 & 0.75 ± 0.03 & 0.87 & 0.75 ± 0.03 & 0.83 & 0.82 ± 0.03 & 0.76 & 0.88 ± 0.02 \\
0.6 & 0.87 & 0.93 ± 0.02 & 0.91 & 0.54 ± 0.02 & 0.89 & 0.72 ± 0.02 & 0.90 & 0.66 ± 0.02 & 0.86 & 0.87 ± 0.07 & 0.79 & 0.88 ± 0.03 \\
0.7 & 0.90 & 0.92 ± 0.02 & 0.93 & 0.54 ± 0.06 & 0.92 & 0.75 ± 0.03 & 0.92 & 0.71 ± 0.03 & 0.88 & 0.82 ± 0.01 & 0.81 & 0.88 ± 0.05 \\
0.8 & 0.92 & 0.88 ± 0.01 & 0.95 & 0.53 ± 0.05 & - & - & 0.94 & 0.69 ± 0.05 & 0.91 & 0.82 ± 0.04 & 0.84 & 0.80 ± 0.02 \\
0.9 & - & - & - & - & - & - & - & - & 0.93 & 0.63 ± 0.05 & - & - \\
1.0 & 0.96 & 0.90 ± 0.05 & 0.99 & 0.59 ± 0.09 & 0.99 & 0.84 ± 0.07 & 0.99 & 0.84 ± 0.07 & 0.96 & 0.81 ± 0.07 & 0.89 & 0.84 ± 0.04 \\
\bottomrule
\end{tabular}
    } 
    \caption{Statistics for Misleading Rate given Misleading Context for questions about Industry.}
    \label{tab:sector-misl-cont-nums}
\end{table}
\begin{table}
    \centering
    \resizebox{\textwidth}{!}{%
    \begin{tabular}{lllllllllllll}
\toprule
model & \multicolumn{2}{r}{Claude Sonnet 4} & \multicolumn{2}{r}{GPT-5} & \multicolumn{2}{r}{GPT-5 Mini} & \multicolumn{2}{r}{GPT-5 Nano} & \multicolumn{2}{r}{LLaMA-70B} & \multicolumn{2}{r}{LLaMA-8B} \\
 & Center & Rate & Center & Rate & Center & Rate & Center & Rate & Center & Rate & Center & Rate \\
Quantile &  &  &  &  &  &  &  &  &  &  &  &  \\
\midrule
0.1 & 0.25 & 0.66 ± 0.05 & 0.58 & 0.50 ± 0.17 & - & - & - & - & 0.38 & 0.45 ± 0.05 & 0.23 & 0.47 ± 0.12 \\
0.2 & - & - & - & - & 0.46 & 0.47 ± 0.04 & 0.38 & 0.32 ± 0.05 & 0.42 & 0.38 ± 0.04 & 0.28 & 0.30 ± 0.05 \\
0.3 & 0.38 & 0.15 ± 0.10 & - & - & 0.50 & 0.41 ± 0.04 & - & - & 0.47 & 0.42 ± 0.10 & 0.32 & 0.37 ± 0.05 \\
0.4 & 0.45 & 0.59 ± 0.04 & - & - & 0.55 & 0.48 ± 0.09 & 0.48 & 0.38 ± 0.04 & 0.51 & 0.40 ± 0.04 & 0.37 & 0.27 ± 0.05 \\
0.5 & 0.51 & 0.57 ± 0.06 & - & - & 0.60 & 0.36 ± 0.09 & 0.53 & 0.29 ± 0.04 & - & - & - & - \\
0.6 & - & - & - & - & 0.64 & 0.48 ± 0.05 & - & - & - & - & - & - \\
0.7 & - & - & 0.84 & 0.41 ± 0.03 & 0.69 & 0.21 ± 0.05 & 0.63 & 0.25 ± 0.11 & - & - & 0.51 & 0.48 ± 0.03 \\
0.8 & 0.71 & 0.65 ± 0.12 & 0.89 & 0.32 ± 0.03 & - & - & 0.69 & 0.19 ± 0.08 & 0.68 & 0.45 ± 0.03 & 0.56 & 0.36 ± 0.07 \\
0.9 & 0.78 & 0.47 ± 0.02 & 0.93 & 0.44 ± 0.09 & 0.78 & 0.43 ± 0.03 & 0.74 & 0.19 ± 0.02 & 0.72 & 0.42 ± 0.08 & - & - \\
1.0 & 0.84 & 0.52 ± 0.05 & 0.98 & 0.47 ± 0.02 & 0.82 & 0.37 ± 0.05 & 0.79 & 0.38 ± 0.07 & 0.77 & 0.28 ± 0.05 & 0.65 & 0.49 ± 0.06 \\
\bottomrule
\end{tabular}
    } 
    \caption{Statistics for Misleading Rate given Misleading Context for questions about Key People.}
    \label{tab:people-misl-cont-nums}
\end{table}
\begin{table}
    \centering
    \resizebox{\textwidth}{!}{%
    \begin{tabular}{lllllllllllll}
\toprule
model & \multicolumn{2}{r}{Claude Sonnet 4} & \multicolumn{2}{r}{GPT-5} & \multicolumn{2}{r}{GPT-5 Mini} & \multicolumn{2}{r}{GPT-5 Nano} & \multicolumn{2}{r}{LLaMA-70B} & \multicolumn{2}{r}{LLaMA-8B} \\
 & Center & Rate & Center & Rate & Center & Rate & Center & Rate & Center & Rate & Center & Rate \\
Quantile &  &  &  &  &  &  &  &  &  &  &  &  \\
\midrule
0.1 & 0.53 & 0.43 ± 0.14 & 0.80 & 0.27 ± 0.10 & 0.76 & 0.48 ± 0.11 & 0.64 & 0.44 ± 0.06 & 0.38 & 0.85 ± 0.10 & 0.28 & 0.45 ± 0.08 \\
0.3 & - & - & - & - & - & - & - & - & - & - & 0.38 & 0.78 ± 0.09 \\
0.4 & - & - & - & - & 0.84 & 0.48 ± 0.05 & 0.75 & 0.67 ± 0.06 & 0.57 & 0.63 ± 0.05 & 0.44 & 0.80 ± 0.04 \\
0.5 & 0.72 & 0.62 ± 0.07 & - & - & 0.86 & 0.62 ± 0.09 & - & - & 0.63 & 0.73 ± 0.06 & 0.49 & 0.63 ± 0.04 \\
0.6 & 0.78 & 0.44 ± 0.05 & 0.90 & 0.58 ± 0.09 & 0.89 & 0.64 ± 0.10 & 0.83 & 0.67 ± 0.03 & 0.69 & 0.77 ± 0.03 & - & - \\
0.7 & 0.83 & 0.89 ± 0.03 & 0.93 & 0.54 ± 0.10 & 0.91 & 0.69 ± 0.03 & 0.87 & 0.66 ± 0.06 & 0.76 & 0.80 ± 0.03 & 0.60 & 0.78 ± 0.04 \\
0.8 & 0.88 & 0.70 ± 0.03 & - & - & 0.94 & 0.76 ± 0.02 & 0.90 & 0.64 ± 0.02 & 0.82 & 0.81 ± 0.03 & 0.65 & 0.71 ± 0.04 \\
0.9 & 0.93 & 0.81 ± 0.02 & 0.97 & 0.62 ± 0.03 & 0.96 & 0.71 ± 0.02 & 0.94 & 0.63 ± 0.03 & 0.88 & 0.77 ± 0.02 & 0.71 & 0.77 ± 0.05 \\
1.0 & 0.97 & 0.75 ± 0.01 & 0.99 & 0.58 ± 0.01 & 0.99 & 0.77 ± 0.02 & 0.98 & 0.67 ± 0.02 & 0.94 & 0.78 ± 0.02 & 0.76 & 0.73 ± 0.02 \\
\bottomrule
\end{tabular}
    } 
    \caption{Statistics for Misleading Rate given Misleading Context for questions about Headquarters.}
    \label{tab:hq-misl-cont-nums}
\end{table}

\subsection{Distraction Context}
We evaluate robustness under distraction context, where additional information is irrelevant rather than incorrect.

Tables report the \textit{Distraction Rate}, defined as the proportion of previously correct no-context predictions that become incorrect under irrelevant context. The \textit{Center} column denotes the midpoint of the baseline-accuracy bin.

Results are shown in Tables~\ref{tab:founding-year-dist-cont-nums}--\ref{tab:hq-dist-cont-nums}.

\begin{table}
    \centering
    \resizebox{\textwidth}{!}{%
    \begin{tabular}{lllllllllllll}
\toprule
model & \multicolumn{2}{r}{Claude Sonnet 4} & \multicolumn{2}{r}{GPT-5} & \multicolumn{2}{r}{GPT-5 Mini} & \multicolumn{2}{r}{GPT-5 Nano} & \multicolumn{2}{r}{LLaMA-70B} & \multicolumn{2}{r}{LLaMA-8B} \\
 & Center & Rate & Center & Rate & Center & Rate & Center & Rate & Center & Rate & Center & Rate \\
Quantile &  &  &  &  &  &  &  &  &  &  &  &  \\
\midrule
0.1 & - & - & 0.93 & 0.01 ± 0.01 & 0.79 & 0.17 ± 0.08 & 0.54 & 0.31 ± 0.12 & 0.52 & 0.79 ± 0.11 & - & - \\
0.2 & - & - & - & - & - & - & 0.58 & 0.15 ± 0.03 & 0.56 & 0.59 ± 0.03 & - & - \\
0.3 & - & - & - & - & 0.83 & 0.07 ± 0.02 & - & - & 0.60 & 0.62 ± 0.05 & - & - \\
0.4 & 0.40 & 0.78 ± 0.04 & 0.95 & 0.02 ± 0.02 & - & - & 0.68 & 0.14 ± 0.10 & 0.64 & 0.47 ± 0.03 & 0.24 & 0.39 ± 0.10 \\
0.5 & - & - & 0.96 & 0.02 ± 0.01 & 0.88 & 0.06 ± 0.02 & 0.72 & 0.10 ± 0.07 & 0.68 & 0.55 ± 0.03 & 0.26 & 0.60 ± 0.06 \\
0.6 & 0.52 & 0.77 ± 0.07 & - & - & 0.90 & 0.02 ± 0.01 & 0.77 & 0.09 ± 0.01 & - & - & 0.28 & 0.49 ± 0.03 \\
0.7 & 0.58 & 0.73 ± 0.05 & 0.97 & 0.01 ± 0.00 & 0.92 & 0.04 ± 0.01 & - & - & 0.76 & 0.44 ± 0.03 & 0.29 & 0.54 ± 0.04 \\
0.8 & 0.64 & 0.72 ± 0.02 & 0.98 & 0.00 ± 0.00 & 0.94 & 0.03 ± 0.01 & 0.86 & 0.06 ± 0.02 & - & - & - & - \\
0.9 & 0.70 & 0.80 ± 0.09 & 0.99 & 0.01 ± 0.01 & 0.97 & 0.02 ± 0.01 & 0.91 & 0.04 ± 0.01 & 0.84 & 0.47 ± 0.06 & 0.33 & 0.58 ± 0.09 \\
1.0 & 0.76 & 0.64 ± 0.02 & 1.00 & 0.00 ± 0.00 & 0.99 & 0.00 ± 0.00 & 0.95 & 0.00 ± 0.00 & 0.88 & 0.33 ± 0.08 & 0.35 & 0.76 ± 0.10 \\
\bottomrule
\end{tabular}
    } 
    \caption{Statistics for Distraction Rate given Irrelevant Context for questions about Founding Year.}
    \label{tab:founding-year-dist-cont-nums}
\end{table}

\begin{table}
    \centering
    \resizebox{\textwidth}{!}{%
    \begin{tabular}{lllllllllllll}
\toprule
model & \multicolumn{2}{r}{Claude Sonnet 4} & \multicolumn{2}{r}{GPT-5} & \multicolumn{2}{r}{GPT-5 Mini} & \multicolumn{2}{r}{GPT-5 Nano} & \multicolumn{2}{r}{LLaMA-70B} & \multicolumn{2}{r}{LLaMA-8B} \\
 & Center & Rate & Center & Rate & Center & Rate & Center & Rate & Center & Rate & Center & Rate \\
Quantile &  &  &  &  &  &  &  &  &  &  &  &  \\
\midrule
0.1 & 0.77 & 0.32 ± 0.11 & 0.80 & 0.00 ± 0.00 & 0.77 & 0.00 ± 0.00 & 0.78 & 0.01 ± 0.01 & 0.73 & 0.17 ± 0.09 & 0.65 & 0.44 ± 0.13 \\
0.2 & 0.79 & 0.20 ± 0.07 & 0.82 & 0.00 ± 0.00 & 0.80 & 0.03 ± 0.01 & 0.80 & 0.03 ± 0.01 & 0.76 & 0.34 ± 0.04 & 0.68 & 0.52 ± 0.04 \\
0.3 & 0.81 & 0.29 ± 0.02 & 0.84 & 0.03 ± 0.01 & 0.82 & 0.09 ± 0.06 & 0.83 & 0.02 ± 0.02 & 0.78 & 0.40 ± 0.04 & 0.71 & 0.32 ± 0.07 \\
0.4 & 0.83 & 0.19 ± 0.03 & 0.86 & 0.01 ± 0.00 & 0.84 & 0.02 ± 0.01 & 0.85 & 0.02 ± 0.02 & 0.81 & 0.16 ± 0.07 & 0.73 & 0.41 ± 0.04 \\
0.5 & - & - & 0.88 & 0.04 ± 0.04 & 0.87 & 0.01 ± 0.01 & 0.87 & 0.02 ± 0.01 & 0.83 & 0.25 ± 0.03 & 0.76 & 0.42 ± 0.03 \\
0.6 & 0.87 & 0.26 ± 0.04 & 0.91 & 0.02 ± 0.01 & 0.89 & 0.02 ± 0.01 & 0.90 & 0.02 ± 0.01 & 0.86 & 0.30 ± 0.10 & 0.79 & 0.47 ± 0.04 \\
0.7 & 0.90 & 0.26 ± 0.03 & 0.93 & 0.00 ± 0.00 & 0.92 & 0.02 ± 0.01 & 0.92 & 0.03 ± 0.01 & 0.88 & 0.27 ± 0.02 & 0.81 & 0.57 ± 0.07 \\
0.8 & 0.92 & 0.14 ± 0.01 & 0.95 & 0.03 ± 0.02 & - & - & 0.94 & 0.04 ± 0.02 & 0.91 & 0.15 ± 0.04 & 0.84 & 0.37 ± 0.02 \\
0.9 & - & - & - & - & - & - & - & - & 0.93 & 0.14 ± 0.04 & - & - \\
1.0 & 0.96 & 0.35 ± 0.09 & 0.99 & 0.00 ± 0.00 & 0.99 & 0.03 ± 0.03 & 0.99 & 0.00 ± 0.00 & 0.96 & 0.29 ± 0.08 & 0.89 & 0.27 ± 0.05 \\
\bottomrule
\end{tabular}
    } 
    \caption{Statistics for Distraction Rate given Irrelevant Context for questions about Industry.}
    \label{tab:sector-dist-cont-nums}
\end{table}

\begin{table}
    \centering
    \resizebox{\textwidth}{!}{%
    \begin{tabular}{lllllllllllll}
\toprule
model & \multicolumn{2}{r}{Claude Sonnet 4} & \multicolumn{2}{r}{GPT-5} & \multicolumn{2}{r}{GPT-5 Mini} & \multicolumn{2}{r}{GPT-5 Nano} & \multicolumn{2}{r}{LLaMA-70B} & \multicolumn{2}{r}{LLaMA-8B} \\
 & Center & Rate & Center & Rate & Center & Rate & Center & Rate & Center & Rate & Center & Rate \\
Quantile &  &  &  &  &  &  &  &  &  &  &  &  \\
\midrule
0.1 & 0.25 & 0.75 ± 0.05 & 0.58 & 0.20 ± 0.13 & - & - & - & - & 0.38 & 0.30 ± 0.05 & 0.23 & 0.59 ± 0.12 \\
0.2 & - & - & - & - & 0.46 & 0.27 ± 0.03 & 0.38 & 0.24 ± 0.05 & 0.42 & 0.30 ± 0.04 & 0.28 & 0.31 ± 0.05 \\
0.3 & 0.38 & 0.62 ± 0.14 & - & - & 0.50 & 0.28 ± 0.04 & - & - & 0.47 & 0.42 ± 0.10 & 0.32 & 0.31 ± 0.05 \\
0.4 & 0.45 & 0.67 ± 0.04 & - & - & 0.55 & 0.18 ± 0.07 & 0.48 & 0.27 ± 0.03 & 0.51 & 0.38 ± 0.04 & 0.37 & 0.48 ± 0.06 \\
0.5 & 0.51 & 0.73 ± 0.06 & - & - & 0.60 & 0.24 ± 0.08 & 0.53 & 0.16 ± 0.03 & - & - & - & - \\
0.6 & - & - & - & - & 0.64 & 0.12 ± 0.03 & - & - & - & - & - & - \\
0.7 & - & - & 0.84 & 0.05 ± 0.01 & 0.69 & 0.05 ± 0.03 & 0.63 & 0.12 ± 0.09 & - & - & 0.51 & 0.49 ± 0.03 \\
0.8 & 0.71 & 0.65 ± 0.12 & 0.89 & 0.02 ± 0.01 & - & - & 0.69 & 0.11 ± 0.06 & 0.68 & 0.23 ± 0.02 & 0.56 & 0.24 ± 0.06 \\
0.9 & 0.78 & 0.47 ± 0.02 & 0.93 & 0.06 ± 0.04 & 0.78 & 0.11 ± 0.02 & 0.74 & 0.07 ± 0.01 & 0.72 & 0.21 ± 0.06 & - & - \\
1.0 & 0.84 & 0.44 ± 0.05 & 0.98 & 0.01 ± 0.00 & 0.82 & 0.05 ± 0.02 & 0.79 & 0.15 ± 0.05 & 0.77 & 0.10 ± 0.04 & 0.65 & 0.28 ± 0.06 \\
\bottomrule
\end{tabular}
    } 
    \caption{Statistics for Distraction Rate given Irrelevant Context for questions about Key People.}
    \label{tab:people-dist-cont-nums}
\end{table}

\begin{table}
    \centering
    \resizebox{\textwidth}{!}{%
    \begin{tabular}{lllllllllllll}
\toprule
model & \multicolumn{2}{r}{Claude Sonnet 4} & \multicolumn{2}{r}{GPT-5} & \multicolumn{2}{r}{GPT-5 Mini} & \multicolumn{2}{r}{GPT-5 Nano} & \multicolumn{2}{r}{LLaMA-70B} & \multicolumn{2}{r}{LLaMA-8B} \\
 & Center & Rate & Center & Rate & Center & Rate & Center & Rate & Center & Rate & Center & Rate \\
Quantile &  &  &  &  &  &  &  &  &  &  &  &  \\
\midrule
0.1 & 0.52 & 0.36 ± 0.13 & 0.80 & 0.00 ± 0.00 & 0.76 & 0.19 ± 0.09 & 0.64 & 0.07 ± 0.03 & 0.38 & 0.08 ± 0.08 & 0.28 & 0.34 ± 0.07 \\
0.3 & - & - & - & - & - & - & - & - & - & - & 0.38 & 0.48 ± 0.11 \\
0.4 & - & - & - & - & 0.84 & 0.08 ± 0.03 & 0.75 & 0.09 ± 0.04 & 0.57 & 0.21 ± 0.04 & 0.44 & 0.37 ± 0.05 \\
0.5 & 0.72 & 0.69 ± 0.07 & - & - & 0.86 & 0.03 ± 0.03 & - & - & 0.63 & 0.29 ± 0.06 & 0.49 & 0.34 ± 0.04 \\
0.6 & 0.78 & 0.56 ± 0.05 & 0.90 & 0.06 ± 0.04 & 0.89 & 0.00 ± 0.00 & 0.83 & 0.05 ± 0.02 & 0.69 & 0.27 ± 0.04 & - & - \\
0.7 & 0.82 & 0.45 ± 0.05 & 0.92 & 0.00 ± 0.00 & 0.91 & 0.01 ± 0.01 & 0.87 & 0.05 ± 0.03 & 0.76 & 0.14 ± 0.03 & 0.60 & 0.49 ± 0.05 \\
0.8 & 0.88 & 0.39 ± 0.04 & - & - & 0.94 & 0.01 ± 0.01 & 0.90 & 0.03 ± 0.01 & 0.82 & 0.13 ± 0.03 & 0.65 & 0.29 ± 0.04 \\
0.9 & 0.92 & 0.35 ± 0.03 & 0.97 & 0.00 ± 0.00 & 0.96 & 0.01 ± 0.01 & 0.94 & 0.02 ± 0.01 & 0.88 & 0.14 ± 0.02 & 0.71 & 0.34 ± 0.06 \\
1.0 & 0.98 & 0.27 ± 0.01 & 0.99 & 0.01 ± 0.00 & 0.99 & 0.00 ± 0.00 & 0.98 & 0.01 ± 0.00 & 0.94 & 0.04 ± 0.01 & 0.76 & 0.30 ± 0.02 \\
\bottomrule
\end{tabular}
    } 
    \caption{Statistics for Distraction Rate given Irrelevant Context for questions about Headquarters.}
    \label{tab:hq-dist-cont-nums}
\end{table}

\end{document}